\documentclass[11pt]{article}
\usepackage{amsmath,amsfonts,amssymb}
\usepackage{graphicx}
\usepackage[backend=bibtex,sorting=none, doi=false]{biblatex}
\usepackage{xurl}
\usepackage{lineno}
\usepackage{hyperref}
\usepackage[title]{appendix}
\hypersetup{colorlinks=true,breaklinks=true, linkcolor=darkcandyapplered, urlcolor=darkcandyapplered, citecolor=darkcandyapplered}
\usepackage{geometry}
\usepackage{times}
\usepackage{titlesec}
\usepackage{booktabs}
\usepackage[table]{xcolor}
\usepackage{tabularx}
\usepackage{longtable}
\usepackage{adjustbox}
\usepackage{caption}
\usepackage{subcaption}
\usepackage{epstopdf}
\usepackage{tikz}
\usepackage{authblk}
\providecommand{\keywords}[1]
{
  \small	
  \textbf{\textit{Keywords---}} #1
}
\usetikzlibrary{positioning}
\usetikzlibrary{shapes.geometric, arrows.meta, positioning}

\title{Integrating Mechanistic and Data-Driven Models for Neurological Disorders through Differentiable Programming}
\author[1]{Shah Pallav Dhanendrakumar}
\author[1]{Saikat Pal}
\author[1,2]{Sitikantha Roy\thanks{Corresponding Author: sroy@am.iitd.ac.in}}
\affil[1]{Department of Applied Mechanics, Indian Institute of Technology Delhi, New Delhi-110016, India}
\affil[2]{Yardi School of Artificial Intelligence, Indian Institute of Technology Delhi, New Delhi-110016, India}
\DeclareBibliographyCategory{cited}
\AtEveryCitekey{\addtocategory{cited}{\thefield{entrykey}}}
\date{}
\definecolor{darkcandyapplered}{rgb}{0.64, 0.0, 0.0}
\begin{document}
\maketitle
\begin{abstract}
Advances in computational modeling, neuroimaging, and artificial intelligence are revolutionizing the modeling of neurological disorders for improved diagnostics, prognosis, and treatment planning. Mechanistic models provide valuable scientific insight into the disorders, but in practice they are often simplified with assumptions or computationally expensive and slow to solve. However, while purely data driven approaches provide speed and scalability, they require large, high quality data to train and generally suffer from interpretability and generalization issues. This perspective paper presents a structured overview of hybrid modeling strategies, which combine deep learning models with physics based solvers, and are categorized into parallel, series, and parallel-series architectures. Three main approaches that have been emphasized are residual modeling for missing or incomplete physics, Neural Ordinary Differential Equations (NODEs) for continuous time dynamics approximation, and solver in the loop that accelerates traditional solvers with neural approximations. These hybrid models integrate the governing differential equation based formulations and deep learning to characterize the evolution of neurological disorders, and promise advanced personalized neurological modeling. In addition, the study explores and proposes different hybrid configurations to improve diagnosis accuracy, predict disease progression, and inform treatment strategies across a range of neurological disorders. To conclude, hybrid modeling offers a compelling trade off between prediction accuracy, computational efficiency, speed and scientific interpretability. These capabilities outperform standalone mechanistic or purely data driven approaches, making hybrid modeling a powerful tool, especially in applications involving modeling the progression and treatment responses in neurological conditions such as brain tumors, Alzheimer’s disease, and stroke.
\end{abstract}
\section{Introduction}
Neurological disorders are a major and increasing challenge to global health, particularly in the context of modern life with its increasing mental workload and stress. The brain is the key organ that controls physiological processes, cognition, and behavior and is highly susceptible to changes in structure and function. Even small abnormalities can cause widespread disruption of many body systems. Such sensitivity highlights the importance of sustained systematic research for a better understanding of the mechanisms underlying normal brain function and disease \cite{BUDDAY2023}. In this context, brain biomechanics has emerged as an important research area providing a physical and mechanical basis to understand how forces, tissue properties, and structural organization influence neurological health \cite{Budday}. The biomechanical approach, unlike the biochemical or genetic approach, highlights the importance of tissue deformation, intracranial pressure, cerebral blood flow, and neural connectivity in the pathogenesis and progression of disease. These are especially relevant in conditions such as hydrocephalus, traumatic brain injury, cerebral aneurysms and stroke \cite{Falaq2020}. Studying these disorders in a mechanical modeling framework allows one to go beyond descriptive observations and develop predictive models to explain injury progression \cite{awasthi2024review}, differences in patient outcomes, and variations in treatment response \cite{Corina2019, awasthi2020biomechanical}. As a result, brain biomechanics provides a strong foundation to improve diagnosis, support clinical decision making, and guide long term disease management.

Alongside these developments, the increasing availability of large clinical, imaging and physiological datasets has led to a greater use of data driven modeling approaches in neuroscience and biomedical research. Such approaches have shown promise for predicting disease progression, treatment effectiveness, and aiding with clinical trial design. Data driven models provide a practical way to describe the complex behavior of the brain by finding patterns in the high dimensional and often noisy data. However, the models based on the statistical or machine learning techniques tend to lack interpretability and generalizability. This limitation is mostly because these models do not contain explicit knowledge of the physical and biological mechanisms. To overcome this problem, recently researchers have focused on combining data driven methods with first principle models, leading to the development of scientific machine learning. \cite{Thompson1994, FORSSELL1997, article, Heiden2021NEURALSIM, Ribba2024, holl2024solving}.

In this context, Physics Informed Neural Networks (PINNs) \cite{Raissi2019} have gained considerable popularity, especially for biomechanical applications \cite{sarabian2021, Sotero2024}. PINNs leverage physical principles represented in the form of differential equations in the training process by adding a residual loss. It ensures that the trained model predictions are consistent with known physics, which is crucial in neurological disease modeling since the data in such cases tend to be limited and noisier \cite{Sotero2024, Sotero2025} than in many other domains. There are several benefits to using PINNs, such as automatic differentiation for gradient calculations, flexible problem solving ability, and elimination of mesh based discretization. On the downside, there are certain obstacles related to computational costs, difficult convergence when dealing with complicated systems, and poor compatibility with existing numerical solvers \cite{Thuerey2021PhysicsBasedDL}. An alternative approach referred to as hybrid modeling overcome these difficulties by embedding the classical numerical solver directly into the machine learning process \cite{SCHWEIDTMANN2024}. Such an approach provides a greater level of control over the precision of numerical calculations, efficient sensitivity analysis, and better scalability for  large and complex models \cite{hernandez2020differentiableprogrammingapplicationsdynamical, sapienza2024}. However, hybrid modeling demands the careful choice of the numerical solver and discretization schemes \cite{ramsundar2021differentiablephysicspositionpiece}. These considerations have increased the interest in hybrid modeling as a more reliable framework for modeling neurological diseases.

A detailed understanding of brain dynamics, especially in diseased states, should combine aspects of data driven adaptive modeling and mechanistic understanding \cite{ramezanian2022generative}. In order to achieve this, hybrid modeling represents the ideal tool for combining deep learning algorithms and differential equation solver together, which can be used for discovering patterns from data and learning physics in a unified manner \cite{Bhutani2006, KANEKO2022}. Using hybrid modeling in brain biomechanics, one can create a model of deformation of tissues, fluids, and neural behavior. As a result, knowledge can be gained about diseases such as brain tumors, neurodegenerative diseases, and stroke. The key feature of the hybrid modeling is its ability to generate predictive, physically consistent, and explainable models while increasing computational efficiency \cite{holl2022physical}. Within this field, there are three prominent approaches: the residual learning approach, neural ordinary differential equations, and the solver in the loop strategy. Each approach has its own advantages and limitations, but together they demonstrate the significance of advancing neurological disorder modeling.

In this study, we present a structured overview of hybrid modeling approaches that integrate deep learning techniques with biomechanical modeling to describe the complex behavior of neurological disorders. The organization of the manuscript is as follows. Section-2 describes the basic concepts required for hybrid modeling, such as differentiable programming, automatic differentiation and adjoint State methods for dynamic systems. We discuss the three principal hybrid model architectures, namely, parallel, series and parallel-series architectures, which combine mechanistic solvers with learning based components, in Section-3. The training strategies and practical benefits and implementation challenges of each architecture are discussed in detail. Section-4 shows how a mechanistic solver can be made differentiable through a simple illustrative example of tumor growth model. Section-5 discusses some selected applications in the domain of modeling of neurological disorders. The focus is on opportunities in tumor growth and treatment planning, progression of neurodegenerative disorders, cerebrovascular modeling, and neuroscience research. Finally, Section-6 concludes the paper with a summary and future directions.

\section{Foundation of Differentiable Physics}
Differentiable physics is an important step forward in computational science because it changes the way physical systems are modeled and optimized. At its core is the idea of differentiable programming for physical systems, which generalizes the end to end differentiability of modern deep learning to the complex mechanistic models that govern real dynamics. The automatic differentiation is a core enabling technology of this approach, which calculates the gradients efficiently through the simulation steps, and avoids the disadvantages of the classical numerical or symbolic differentiation. From this point of view, differentiable physics combines traditional differential equation solvers with deep learning techniques to enable efficient solutions to inverse problems, data driven discovery of governing equations and parameter tuning in the presence of sparse and noisy data. For fields such as biomechanics and neurological modeling, this framework offers powerful opportunities to create models that are robust, interpretable, and scalable, critical qualities for advancing our understanding of health and disease.
\subsection{Principles of Differentiable Programming for Physical Systems}
For complex dynamical systems, progress beyond conventional computational methods is necessary \cite{willard2020}. The strategy which can be anchored through the differentiable programming of the physical system abbreviated as Differentiable physics \cite{holl2024solving} will require combining data driven component and the traditional mechanistic simulation approach together. Differentiable Physics is a robust and flexible framework for developing sophisticated and data informed models.

The main idea of Differentiable physics is to keep all calculations of the simulation of any dynamical system in a form of a computational graph and use this for calculating gradients of any intermediate computational step with respect to previous outputs \cite{Naumann2011}. This allows gradients to be computed efficiently and automatically. Unlike traditional numerical differentiation, which involves finite differences and is often computationally costly and numerically unstable for high dimensional parameter spaces \cite{blondel2025}, Differentiable Physics applies the chain rule of calculus. It then walks through the computational graph using a more sophisticated approach called reverse mode automatic differentiation (or backpropagation in the context of machine learning) to precisely compute the gradient of a scalar loss function with respect to each input parameter or intermediate variable. This precision is invaluable for complex physical models \cite{Baydin2018, Griewank2008}.

Neurological disorders can be modeled as a deformation of brain tissue under various physiological or pathological conditions \cite{ciric2024}. Simulating such model traditionally requires to solve the forward model with fixed brain material parameters. The brain material parameters can be estimated through various mechanical testing over cadaveric brain \cite{gautam2025understanding, gautam2026biomechanical, singh2023mechanical}and inverse FEM \cite{awasthi2022material, awasthi2023matnli}. Addressing inverse problems in the field of brain biomechanics to estimate material properties from observed clinical data \cite{awasthi2022comparison}  or ex-vivo / in-vivo experimental data, often presents significant computational complexities \cite{InverseProblemsBook, renner2024, hinrichsen2023inverse}. The primary hurdle in the traditional computation of inverse problems is gradient computation, as the required number of forward simulations to be solved  grows linearly with the number of model parameters. It becomes more computationally expensive if the dynamical system has high degree of freedom. Differentiable Physics elegantly bypasses these challenges by rendering the entire simulation pipeline differentiable. This means that if we can preserve the entire forward simulation with its loss function as a computational graph, differentiable Physics can seamlessly and efficiently compute the gradients of this loss with respect to the underlying physical parameters with a single forward simulation \cite{newbury2024reviewdifferentiablesimulators, Farsi2025}.
\subsection{Automatic Differentiation (AD): Enabling Gradient Based Learning via the Computational Graph}
The remarkable success of deep learning in diverse fields has been largely due to the efficient optimization capabilities offered by gradient based methods, achieved due to backpropagation through complex computational graph. 
The core of this efficiency relies on Automatic Differentiation (AD) \cite{FISCHER1993}, which is a set of techniques that compute exactly the differentials of a function as represented by a computer program. While AD is frequently compared alongside symbolic and numerical differentiation, it enjoys a unique and better position, especially in relation to high complexity and high dimensionality functions as is the case in physical simulations \cite{Griewank2008}. For researchers using computer science to address complex problems in biomechanics and modeling the neurological conditions, mastering AD is crucial for leveraging the full capabilities of differentiable physics.

The general principle of AD is that any computer program, no matter how complicated, is just a sequence of elementary arithmetic operations (like addition, multiplication) and elementary functions (like sine, cosine, exponential). AD computes the derivatives of the program output with respect to the provided inputs exactly by systematically applying the chain rule of calculus to these elementary operations \cite{fang2024ADImplementation}. AD computes derivatives to machine precision at a computational cost that grows linearly with the cost of evaluating the original function \cite{Baydin2018}. This efficiency is important to train models for complex physical simulations. The importance of AD in modern deep learning frameworks (e.g., TensorFlow, PyTorch, JAX) can be seen clearly as the usage by the scientific community is growing rapidly. Two main modes of AD are widely employed:
\begin{enumerate}
    \item \textbf{Forward Mode AD:} This mode propagates derivatives forward through the computational graph. For each elementary operation, it computes both the function value and its derivative at node  by leveraging dual number structure and push them in forward direction. 
    As it calculates the gradient for one input variable in a single pass, it requires exactly the same number of passes to the number of inputs. Although easy to implement, its computational cost scales linearly with the number of input variables if we need gradients for all inputs, making it less efficient for functions with many inputs and few outputs \cite{Baydin2018}.
    \item \textbf{Reverse Mode AD (Backpropagation):} This mode computes the functional value by propagating forward through the computational graph. To calculate the gradient, the adjoint flows backward from the output to the inputs through all intermediate nodes. This mode is efficient for models with large number of input parameters compared to number of model outputs, which is usually the case in dynamic system modeling with a scalar loss function. In contrast to forward mode AD, in only one pass it can provide gradient with respect to all inputs together, makes it computationally cost effective and suitable choice for dynamic system modeling \cite{Rumelhart1986, werbos1994}. The backpropagation algorithm, now ubiquitous in deep learning, is a specific application of reverse mode AD.
\end{enumerate}
\subsection{Adjoint State Method (ASM): Enabling Gradient Based Learning via Classical Solvers}
In the context of classical solvers, ASM is the enabling technology for gradient based learning. For instance, in a biomechanical simulation of brain deformation, the simulation code itself can be viewed as a function that takes physical parameters (e.g., brain tissue material parameters, boundary conditions) as inputs and produces observable quantities (e.g., displacement fields, stress distributions) as outputs. By applying ASM, we can automatically obtain the gradients of a defined loss function  with respect to these underlying physical parameters without requiring access to the entire computational graph. Furthermore, the computational graph of unrolled iterative solver can be bypassed utilizing ASM. This capability allows for efficient inverse problem computation, which is highly essential in the field of biomechanics.
\begin{figure}[h!]
    \centering
    \includegraphics[width=0.7\linewidth, trim={16cm 8cm 2cm 7.5cm}, clip]{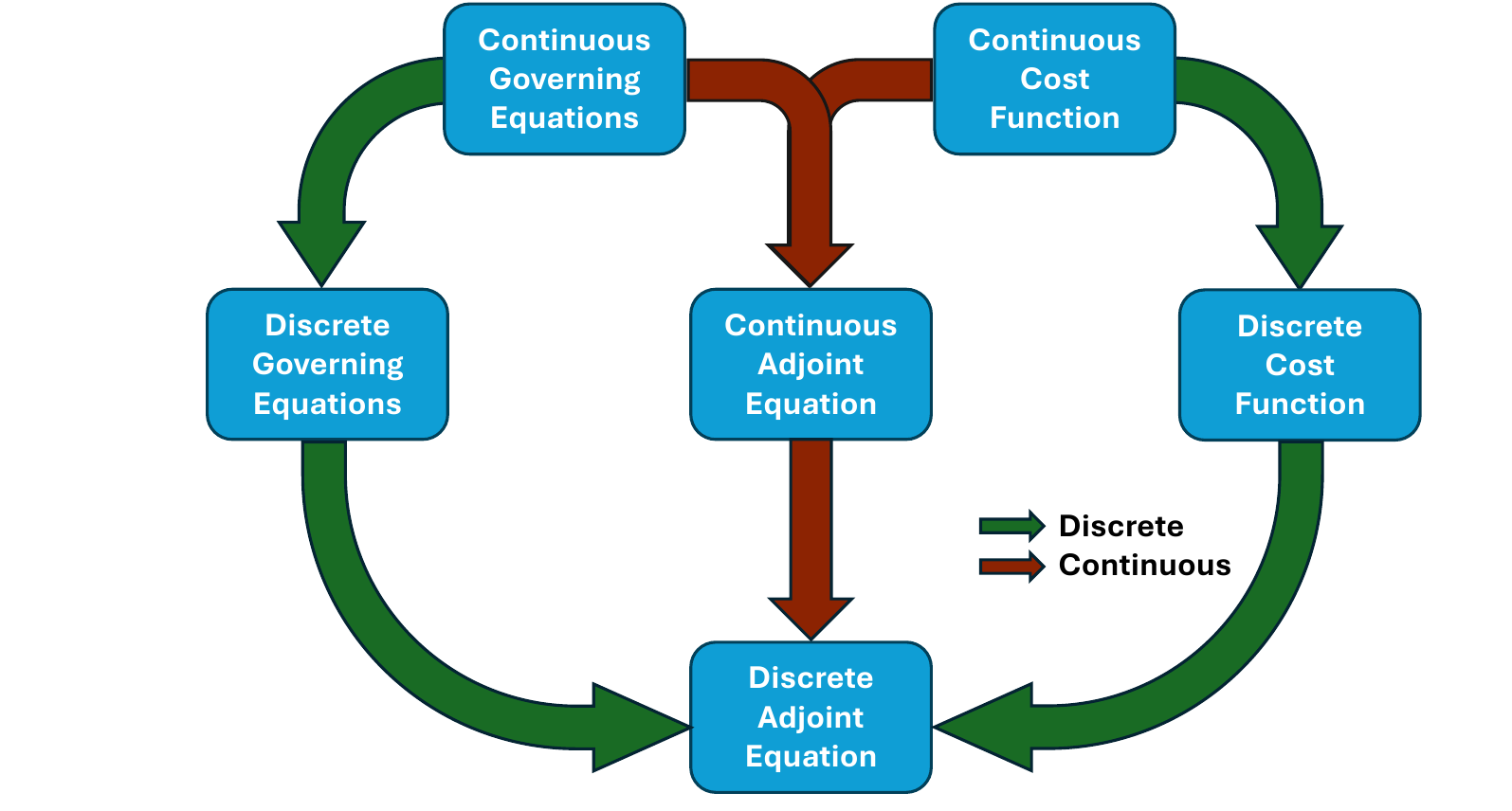}
    \caption{Different approaches of adjoint state method \cite{hekmat2016ASMcomparison}}
    \label{adjoint equation}
\end{figure}
Theoretically, the ASM can be implemented in two different ways known as continuous ASM (Optimize then Discritise - OtD) and discrete ASM (Discritise then Optimize - DtO). The theoretical workflow for arriving at discrete adjoint equation in both methods can be observed in Figure \ref{adjoint equation}. The detailed step by step implementation is described in Figure in the Appendix for an arbitrary dynamic system in general. In view of application to the brain dynamical model, the implementation of ASM has been explained for a Mass-Spring-Damper (MSD) system which can model the response to traumatic brain injury \cite{Laksari2020,Wittek2019} or the viscoelastic nature of brain tissues \cite{Hrapko2006,Rooij2016} as delineated in Figure in the Appendix. Christopher Rackauckas et al. \cite{DifferentiablePhysicsReview} pointed out that the choice between continuous ASM and discrete ASM for the trade off between stability and efficiency depends on the system dynamics equations and the chosen discretization scheme. 
\begin{table}[h!]
\begin{tabular}{p{0.2\linewidth} p{0.35\linewidth} p{0.35\linewidth}}
\toprule
\textbf{Feature} & \textbf{Automatic Differentiation (For Artificial Neural Network)} & \textbf{Adjoint State Method (for Neural ODEs)} \\
\midrule
\textbf{Primary Goal} & Compute exact gradients of a computational graph. & Compute exact gradients of a loss function through an ODE solution. \\
\textbf{Mechanism} & Direct application of chain rule by storing intermediate values and backtracking. & Solves an auxiliary ODE (the adjoint ODE) backward in time, avoiding storage of full trajectory. \\
\textbf{Applicability} & General for any differentiable program (e.g., standard neural networks). & Specifically designed for differentiable programming over continuous dynamical systems (e.g., ODEs, PDEs). \\
\bottomrule
\end{tabular}
\centering
\caption{Comparison of Automatic Differentiation and Adjoint State Method}
\label{tab:ad_adjoint_comparison}
\end{table}
Although AD enables the computation of derivatives of the system equations, but the adjoint method provides the overarching framework for
organizing and accumulating these derivatives over time to obtain the final gradients efficiently, without storing all intermediate states like standard backpropagation does. For neurological disorder modeling, integrating mechanistic insights is most required for developing robust and reliable models \cite{Raissi2019} that can deploy in standard clinical practice where clinical data is mostly sparse, noisy, and expensive to acquire. More accurate and accelerated model can be emerged by the seamless embedding of classical physical models with deep learning architectures.  In such models, the learning process is guided not only by available data but also by the fundamental physical laws governing the system. This makes it possible to train hybrid models for long and complicated continuous processes, which are common in scientific and engineering applications, such as advanced neurological modeling. In short, the hybrid model takes the standard mechanistic model from being just a way to understand science and purely data driven models from being just ways to make predictions and turns them into an adaptive, trainable optimization framework. This makes it possible to use data to find incomplete physics and make predictions that are consistent with the laws of physics. This method is very important for helping us learn more about complicated biological systems like the brain, where the complicated interactions between neural processes and biomechanical responses are the key to solving important problems in neurological health and disease research \cite{sapienza2024}.
\section{Key Hybrid Model Frameworks: Architectural Configurations, Training Regimen , Advantages, Challenges, and Adaptation Strategies}
Psichogios et al.\cite{Psichogios1992} and Hong-Te Su et al. \cite{SU1992}  initiated the work in the direction of hybrid modeling by introducing the hybrid neural network-first principles model and the integrated neural network (INN) model respectively for dynamic behavior of process systems in chemical engineering to incorporate prior physics knowledge and data driven predictability simultaneously. As first principal models provide scientific basis for understanding the phenomenon, they are known as white box models. On the other end, deep learning models lack interpretability as being purely data driven, they are known as black box models. Several researchers propelled the growth in the direction of high predictability with scientific interpretation and developed hybrid models known as gray box models. Motivated by the individual capabilities of deep learning models and mechanistic models, three prominent hybrid configurations have emerged as \textbf{ Parallel, Series, Parallel-Series}. These configurations offer data pattern identification, universal approximation capabilities, physical consistency, and interpretability all together. Understanding the workings of these hybrid models is of crucial interest to researchers who wish to harness them for the modeling of neurological disorders.
\subsection{Parallel Hybrid Models}
\begin{figure*}[!ht]
    \centering
    \begin{subfigure}[!ht]{0.42\linewidth}
        \centering
        \includegraphics[scale=0.3]{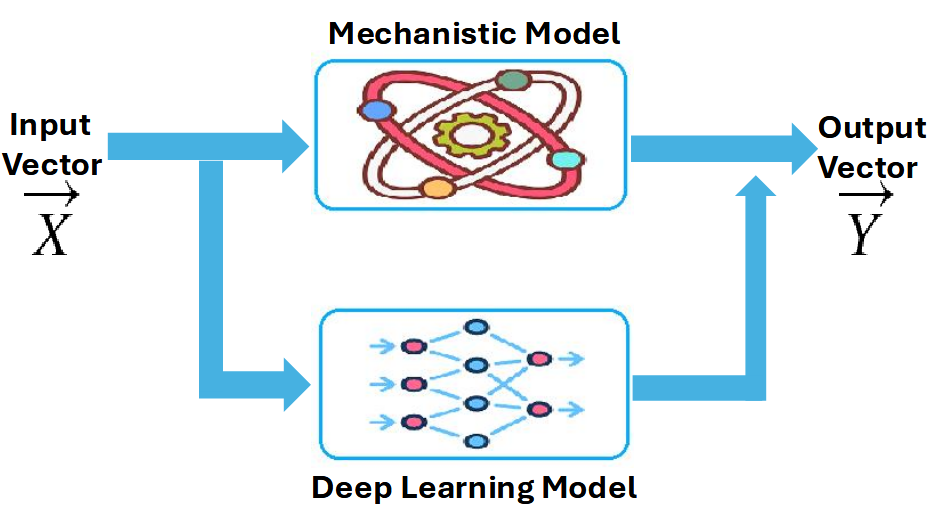}
        \caption{}
    \end{subfigure}%
    \hfill
    \begin{subfigure}[!ht]{0.55\linewidth}
        \centering
        \includegraphics[scale=0.25]{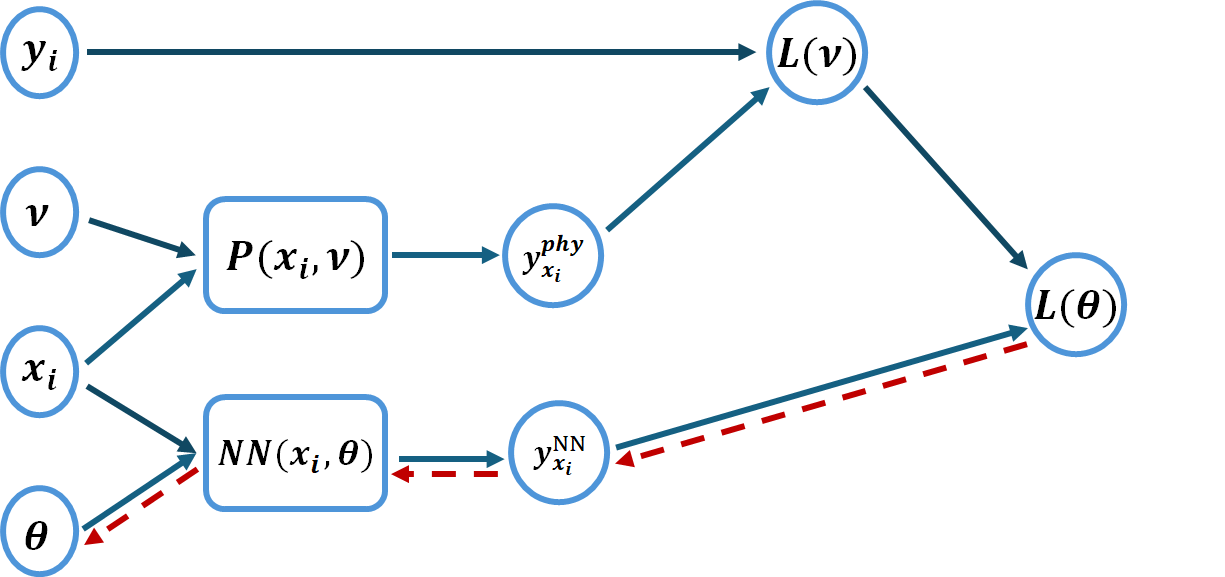}
        \caption{}
    \end{subfigure}%
     \caption{Parallel or Residual Model (a) Architectural Configuration (b) Computational Graph for Training}
     \label{Residual Model}
    \end{figure*}

In a parallel configuration, deep learning models and mechanistic models or underlying classical solvers operate concurrently towards a common objective without entirely dictating the execution flow of each other, as shown in Figure \ref{Residual Model}. The century old efforts of the scientific community have resulted in development of the first principle based mechanistic models which capture a significant amount of system dynamics. Hence, the only objective here is to learn the residual or unknown underlying physics of
the dynamical system from the observation data.\\
\subsubsection{Training Regimen}
To decipher the training regimen of the parallel or residual model, let us focus on the computational graph illustrated in Figure \ref{Residual Model} (b). Let $P(x_i; \nu)$ denote the approximate physical model, parameterized by $\nu$, which predicts an output $y^{Phy}_{x_i}$ for a given input $x_i$. A deep learning model, $NN(x_i; \theta)$, with parameters $\theta$, is then augmented to accurately predict the residual $y^{NN}_{x_i}$. The training procedure unfolds in following steps.
\begin{enumerate}
    \item[TS-1] \textbf{Residual Target Generation} The mechanistic parameter $\nu$ is fitted through the well established scientific approach to capture the significant amount of physics behind system dynamics under consideration. For each input data $x_i$ and fixed $\nu$, the physical model $P(x_i; \nu)$ produces its approximate output $y^{Phy}_{x_i}$ the corresponding true observations $y_i$. The loss of the mechanistic model $L(x_i, \nu) = (y_i - y^{Phy}_{x_i})$ is calculated not to optimize the mechanistic parameter $\nu$. Instead, it is calculated to identify the discrepancy between the observed data $y_i$ and the prediction of mechanistic model $y^{Phy}_{x_i}$, serving as the residual target for optimizing the deep learning model. Once the mechanistic parameter $\nu$ is fixed, it remains unchanged throughout the span.
    \item[TS-2] \textbf{Objective Function Creation} The deep learning model with the learnable parameter $\theta$ is augmented to focus on identifying the missing physics generated here as the residual target. Training of the deep learning model is driven by minimizing a loss function, $L(\theta)$, which quantifies the difference between the predicted residual $y^{NN}_{x_i}$ of the deep learning model and the residual target $L(x_i, \nu)$. For a dataset of $N$ samples, if the mean squared error loss is employed:
    \begin{align}
        L(\theta) = \frac{1}{N} \sum_{i=1}^N \left( y^{NN}_{x_i} - L(x_i, \nu) \right)^2
    \end{align}
    \item[TS-3] \textbf{Gradient based Optimization} As customary in deep learning, the gradient based optimization algorithms (e.g., Adam, Gradient Descent) are utilized for training. Optimization of $L(\theta)$ requires computing the gradient $\frac{\partial L}{\partial \theta}$ via standard backpropagation through the path (red dashed) exhibited in Figure \ref{Residual Model} (b) efficiently using automatic differentiation.
    \begin{align}
        \frac{\partial L}{\partial \theta} = \sum_{i=1}^N \frac{\partial L}{\partial y^{NN}_{x_i}} \frac{\partial y^{NN}_{x_i}}{\partial \theta}
    \end{align}
One can observe the gradient flows backward through deep learning model and not through mechanistic model. However, mechanistic model contribute in training indirectly due to residual target generation $L(x_i, \nu)$as follow:
\begin{align}
    \frac{\partial L}{\partial y^{NN}_{x_i}} &= \frac{2}{N} \left( y^{NN}_{x_i} - L(x_i, \nu) \right)\nonumber\\
    \implies\frac{\partial L}{\partial \theta} &= \sum_{i=1}^N \frac{2}{N} \left( y^{NN}_{x_i} - L(x_i, \nu) \right) \frac{\partial y^{NN}_{x_i}}{\partial \theta}
\end{align}
Training loop initiate with utilizing random value of $\theta$, and an optimizer updates $\theta$ using these gradients to reduce the loss. This process continues for multiple epochs until the deep learning model effectively learns to approximate the residuals, thereby correcting the physical model's inaccuracies.  The system's final prediction for a given $x_i$ would then be $y^{Phy}_{x_i} + y^{NN}_{x_i}$.   
\end{enumerate}
 Universal Differential Equation (UDE) \cite{DifferentiablePhysicsReview, Menard2023ThesisBioprocesses, ElGazzar2024, philipps2025currentUDE} is a significant application of the parallel configuration with model assumption refinement due to which the model is unable to capture a significant amount of system dynamics. In UDE, deep learning model is trained to produce the missing physics as $f_{DL}(t,\vec{X},\theta)$ instead of residual. Here, instead of correcting the output of mechanistic model, the model itself is modified utilizing the deep learning model output.  This hybrid approach offers a robust way to infuse data driven learning into physics-based models, especially valuable when complete physical understanding regarding the underlying mechanism or dynamics is not possible or unknown.
\subsubsection{Advantages}
   On one hand, the mechanistic model enforces first principles knowledge, and on the other hand, deep learning models the residual errors between model predictions and actual observed data. In this way, the model provides adaptability and interpretability together.
\begin{enumerate}
    \item[A1] \textbf{Improved Prediction Accuracy}  The deep learning model augmented here only learn the imperfection of a mechanistic model accounted for complex interactions, random noises, stochastic inferences. This targeted correction strategy accelerates the convergence rate and results in improved prediction accuracy. 
    \item[A2] \textbf{Learning Incomplete Physics} In many cases, the gap between predicted and observed data widens as the mechanistic model simplifies the governing equations and ignores the contribution of secondary processes. The knowledge gap is filled, and reduce the impact of incomplete physics due to the augmentation of deep learning model without discarding the valuable structure of the physical model. 
    \item[A3] \textbf{Computational Efficiency} The training loop does not require the gradient flow through the physical solver, resulting in relatively less computational burden. During optimization, solver do not require repeated evaluations because of not being part of training loop. Therefore, the parallel model offers faster training and wider scalability.
    \item[A4] \textbf{Generalization and Flexibility} The purely data driven approaches often exhibit the overfitting towards training data and struggle to extrapolate accurately; however, the inclusion of mechanistic solvers enforces the physics on extrapolation to mitigate the overfitting risks. As the mechanistic model exists alongside empirical data in diverse scientific and engineering domains, generalization capability and flexibility of the parallel configuration allow seamless adaptation across applications.
\end{enumerate}
\subsubsection{Challenges and Adaptation Strategies}
The parallel configuration, often referred to as residual model \cite{SU1992, Psichogios1992, Oliveira2004} offers computational efficiency since no solver is required inside the training loop, but introduces the following limitations:
\begin{enumerate}
    \item[C1] \textbf{Error accumulation and model bias} If the underlying physics model fails to capture essential mechanisms, the neural correction may converge to biased dynamics, limiting predictive accuracy across operating regimes \cite{Quaghebeur2021, SHAH2025}.
    \item[C2] \textbf{Limited extrapolation capability} Given that residuals are acquired from measured disparities, such parallel models may struggle to generalize without robust physics priors \cite{Rajulapati2022}.\\
    \textbf{Adaptation Strategies}\\ 
Regularization should be enforced by encouraging smoothness that has physical interpretability. Shah et al. \cite{SHAH2025} argue that these residual learners should not be left unconstrained, but instead should be informed by priors in physics knowledge to maintain interpretability and mitigate overfitting.
\end{enumerate}
This parallel structure offers the adaptability and flexibility inherent in deep learning while retaining the interpretability and robustness characteristic of mechanistic models. In terms of brain dynamics, we can utilize this framework by first establishing a baseline physics informed model (i.e. hyperelastic or viscoelastic models with anisotropy) and then using a deep learning model to learn the residual/unknown physics that could be present due to simplifications made in the model (i.e. complex geometrical interactions, tissue-tool interactions that our current mechanistic model does not take into account). By doing this, we are essentially embedding data driven corrections to a physics informed model allowing us to create more accurate and robust neurological models \cite{Raissi2019}.

\subsection{Series Hybrid Models}
Unlike parallel models, serial architectures feature a dependency chain, where one model's output becomes crucial input for another.  This can manifest in several ways, often with the deep learning model either enhancing or being followed by a classical solver for a mechanistic model. One common serial approach involves deep learning models acting as surrogates, preconditioner, or accelerators for classical numerical solvers \cite{AcceleratingSolvers}. For instance, a deep learning model might be trained to provide an initial guess for an iterative solver, significantly reducing convergence time. Alternatively, a deep learning model could learn to accelerate specific computationally intensive steps within a larger classical simulation pipeline, such as solving linear systems or predicting optimal meshing strategies for finite element analysis. This is highly relevant for large scale simulations, where even minor accelerations can translate into substantial computational savings for researchers with limited computational resources. Another powerful serial configuration involves embedding classical solver principles as a differentiable pipeline directly within a differentiable framework to propagate gradients through both the solver and the deep learning parts of the framework \cite{newbury2024reviewdifferentiablesimulators}.
\subsubsection{Neural Ordinary Differential Equations (NODEs)}
\begin{figure*}[!ht]
        \centering
        \begin{subfigure}[!ht]{0.4\linewidth}
        \centering
        \includegraphics[scale=0.35]{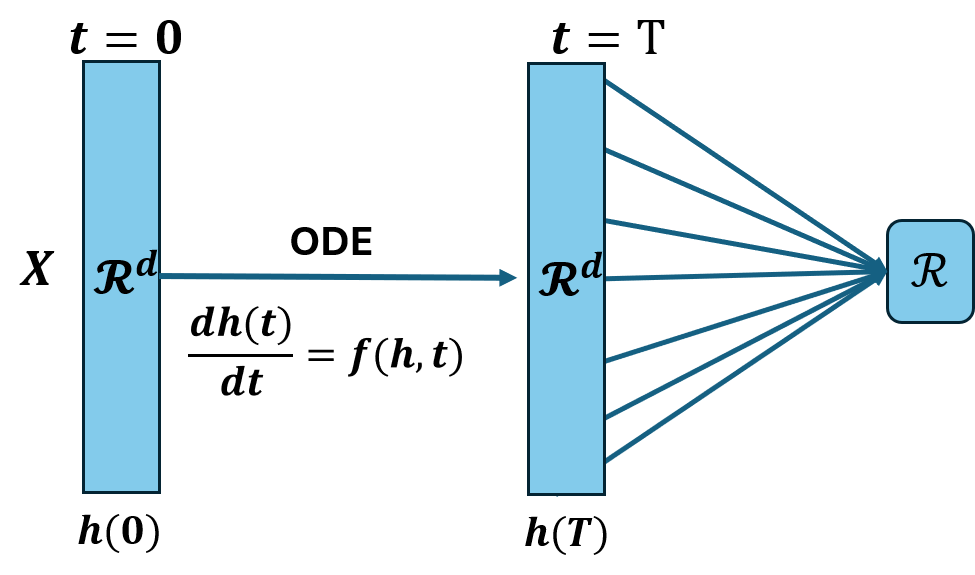}
        \caption{}
    \end{subfigure}%
    \hfill
     \begin{subfigure}[!ht]{0.6\linewidth}
        \centering
        \includegraphics[height=4cm, width=\linewidth]{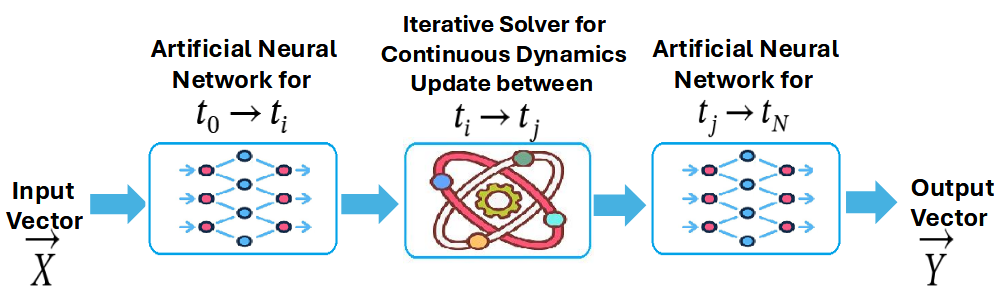}
        \caption{}
    \end{subfigure}%
     \caption{Neural ODEs Architecture (a) Evolution of hidden states governed through the ODE (b) Iterative Solver for the Neural ODE  }
     \label{Neural ODEs Model}
    \end{figure*}
A prominent example of a serial architecture is the \textbf{Neural Ordinary Differential Equation (NODE)} \cite{Chen2018NeuralODE}, where the deep learning model defines the dynamics to be solved by a classical method. NODEs are a family of deep neural network models that can be interpreted as a continuous equivalent of Residual Networks (ResNets).
Unlike traditional ResNets or recurrent neural networks (RNNS) that discretize time into fixed steps, NODEs model continuous time dynamics. Here, a neural network parametrizes the time derivative of the hidden state as shown in Figure \ref{Neural ODEs Model} (a). In the perspective of series configuration, one should not think of replacing the whole update process of discretized hidden states with continuous dynamics. Instead, think of implementing continuous dynamics through an iterative solver to partially replace the update process of discretized hidden states, as delineated in Figure \ref{Neural ODEs Model} (b). The solution to this ODE is then obtained by a standard, off the shelf ODE solver (a classical component), such as the Runge-Kutta method or any adaptive iterative method. 

The key innovation of NODEs lies in their differentiability. The gradients required for training the neural network parameters can be computed by solving an augmented ODE backward in time using the ASM. This allows for efficient backpropagation through the ODE solver, enabling the learning of complex, continuous time dynamics directly from data. In the context of brain dynamics, NODEs offer a powerful way to model the continuous evolution of neural activity, or even the slow, continuous deformation of brain tissue over time under pathological conditions, providing a more natural representation than discrete step models. Their ability to handle variable time steps and potentially fewer parameters than traditional RNNs makes them an attractive option for complex biological systems \cite{oh2025}.
\paragraph{Training Regimen}
Neural ODEs define the transformation of an input as the solution to an ordinary differential equation, where a neural network parameterizes the vector field (the right hand side of the ODE).  The objective when training a Neural ODE is to learn the optimal parameters $\theta$ of the neural network $f$ such that the solution $z(t)$ of the ODE accurately describes observed data or minimizes a specific objective function. The training process conceptually mirrors that of standard neural networks, but with a crucial difference in computing gradients through ASM during the backward pass. Instead of storing the entire forward pass trajectory, the adjoint method introduces an adjoint state (or adjoint variable), which satisfies its own ordinary differential equation.
For Neural ODE, the stepwise ASM procedure is depicted in Figure and training steps are as follow:

\begin{enumerate}
    \item[TS-1] \textbf{Forward Inference} Given an initial state $z(t_0)$ and the current parameters $\theta$ of the neural network $f$, a standard ODE solver (e.g. Runge-Kutta, Adams-Bashforth) is used to solve the ODE from $t_0$ to $t_N$. This integration yields the predicted state $z(t_N)$ (or the complete trajectory $z(t)$ over $[t_0, t_N]$).  During this pass, only a few checkpoints of the state, say $s(t_0),\dots,s(t_i),s(t_{i+1})\dots,s(t_N)$ need to be stored, not the entire high resolution trajectory. This step essentially represents the prediction made by the Neural ODE.
    \item[TS-2] \textbf{Loss Calculation:} A scalar loss function $L$ is then computed, quantifying the discrepancy between the predicted state $s(t_N)$ (or trajectory) and the actual observed target data. 
    \item[TS-3] \textbf{Gradient based Optimization} Calculating the gradient $\frac{\partial L}{\partial \theta}$ is the most critical and distinct process. In standard neural networks, reverse mode AD can be applied directly to the entire network graph for the gradient computation step.  The challenge in training NODEs arises because the layers are not discrete but represent a continuous evolution governed by an ODE. Since forward inference is performed by an ODE solver, a naive application of reverse mode AD would require storing the entire trajectory of the hidden state at every tiny step taken by the ODE solver, leading to prohibitive memory requirements for long integration times or high precision solvers \cite{Chen2018NeuralODE}. This is precisely where the ASM becomes indispensable. The ASM introduces an auxiliary variable, the adjoint $a(t)$ satisfying a backward in time ODE, often referred to as the adjoint ODE:
       \begin{align}
           &\frac{da(t)}{dt} = -a(t)^T \frac{\partial f(z(t), t, \theta)}{\partial z}\nonumber\\
           \hspace{-5cm}\text{with a boundary condition:}\,\,\,\, &a(t_N) = \frac{\partial L}{\partial z(t_N)}
       \end{align} 
        The gradient of the loss with respect to the parameters $\theta$ is then computed by integrating another term involving the adjoint and the partial derivative of the vector field $f$ with respect to $\theta$:
        \begin{align}
            \frac{dL}{d\theta} = \int_{t_0}^{t_N} a(t)^T \frac{\partial f(z(t), t, \theta)}{\partial \theta} dt
        \end{align} 
        This integration can also be formulated as solving an augmented ODE backward in time. Finally, an optimization algorithm uses these computed gradients to iteratively adjust the neural network parameter $\theta$, minimizing the loss function and improving the Neural ODE's ability to model the observed dynamics.
\end{enumerate}  
           
\paragraph{Advantages}
\begin{enumerate}
    \item[A1] \textbf{Lower Memory Requirements} Neural ODE uses ASM to calculate gradients instead of using standard backpropagation. As a result, even for nonlinear problems memory usage grows only linearly, empowering Neural ODE to be trained without excessive memory requirement.  
    \item[A2] \textbf{Accuracy-Efficiency Balance} Adaptive solvers are being used for Neural ODEs, making it possible for researchers to choose suitable step sizes and tolerances as their priority between accuracy and efficiency. This makes Neural ODEs most favorable choice in real time resource limited scenarios. Additionally, Neural ODE provides flexibility as depth is not fixed in adaptive solvers.
    \item[A3] \textbf{Broader Modeling Capabilities} Like machine learning models, Neural ODEs also provide smooth and invertible mapping, tractable change of variables and capability to handle irregularly sampled time series data.
\end{enumerate}
\paragraph{Challenges and Adaptation Strategies}
Neural ODEs exemplify continuous depth models, where trajectories are implicitly defined and approximated by a numerical solver \cite{Chen2018NeuralODE}. Despite their elegance, two major challenges arise:
\begin{enumerate}
    \item[C1] \textbf{Memory limitations} Training NODEs involves differentiation along solver trajectories. During backpropagation, memory consumption mainly depends on the necessity to keep all hidden states in memory across the whole computational path since they will be used for the gradient calculation. This problem is especially pronounced for deep or continuous depth models due to a long computational graph. One possible approach is to store all the solver states that is quite costly in terms of memory requirements. The ASM approach provides an efficient solution by restoring gradients using a backward ODE integration. However, solver approximation often break reversibility, resulting in instability \cite{Chen2018NeuralODE, DeepImplicitLayers}.
    
    \textbf{Adaptation Strategies}\\
     Some commonly used strategies include:
    \begin{itemize}
        \item \textbf{Checkpointing:} 
        T In this strategy, a subset of states referred to as checkpoints are saved while doing a forward pass through the model \cite{DeepImplicitLayers}. Afterward, if any gradients need to be restored, other intermediate state should be calculated starting from the closest checkpoint \cite{Chen2018NeuralODE}. Although the proposed method significantly decreases peak memory usage, it incurs additional computation costs because the same trajectory in fragments have to be executed several times. The efficiency of checkpointing is determined by the frequency of checkpoints which balances the trade off between memory efficiency and numerical stability during backward integration.
        \item \textbf{Reversible architectures:} A different method that allows saving on memory is the application of reversible architectures, in which the requirement for memory space is completely avoided since one is able to reconstruct previous states exactly from latter ones. The reverse time integration of an ODE is equivalent to forward time integration, where the vector field dynamics of the ODE are reversed to $y'= - f(- t,y)$. As it can be observed for ODE-based models, there is another advantage related to reversibility since exact ODE solutions inherent reversibility.
    \end{itemize}
       In practice, however, backpropagation can diverge as a consequence of numerical instabilities, thus leading to the emergence of hybrid and solver aware techniques. In order to obtain smoother dynamics and, therefore, less solving steps, some regularization methods are considered, namely the introduction of penalties related to large norms or hidden state velocities. Mixed strategies, in which checkpointing together with adjoint state integration of small segments \cite{Chen2018NeuralODE} are used, balance computational cost with memory savings and avoid catastrophic gradient errors. This results in efficient implementation at large scale. Together, these strategies extend the practical scalability of Neural ODEs, ensuring that their theoretical memory efficiency can be realized in large scale applications such as neurological disorder modeling.

    \item[C2] \textbf{Expressivity restrictions}
   For NODEs to ensure well posedness of solution, it must assume existence and uniqueness (via Lipschitz continuity), thus penalizing the use of non smooth activations. Additionally, ODE flows are invertible, and thus NODEs can only approximate smooth homeomorphism. The structure of NODEs prevents crossing trajectories \cite{DeepImplicitLayers}, limiting expressivity. NODEs also inherently assume deterministic flows, which may not be ideal if the desired dynamics are noisy or stochastic \cite{Tzen2019NeuralSDEs}.\\ 
    \textbf{Adaptation Strategies}\\
    To extend the applicability of NODEs beyond purely deterministic, smooth and invertible flows, and to restore expressivity for complex real world systems, we update standard NODE with architectural modification tailored to different dynamical contexts. Augmented Neural ODEs \cite{Dupont2019AugmentedNeuralODEs} overcome the smooth homeomorphism restriction by embedding in higher dimensional latent spaces, and second order NODEs and hybrid architectures introduce flexibility. Neural SDEs \cite{Tzen2019NeuralSDEs}, Neural CDEs \cite{Kidger2020NeuralCDEs} generalize NODEs to stochastic or input driven regimes and ensemble methods can provide uncertainty estimates when full stochastic formulations are computationally unfeasible. Another key direction in NODE variants is to incorporate principles from physics and modeling uncertainty. These physics informed variants \cite{cranmer2019lagrangian, Greydanus2019HamiltonianNN, Karn2024} perform better and are more interpretable on physical systems tasks. Collectively, these variants showcase the active research in expanding the theoretical foundations and practical applicability of NODEs \cite{SecondOrderAugmentedODE, irie2022neural, yi2023nmode}, making them applicable to noisy, irregular, or physically constrained neurological systems.
\end{enumerate}
\subsubsection{Solver in the Loop}
The solver in the loop setup integrates a traditional differential equation solver with the deep learning model. The entire process, including the solver's internal computations, must be fully differentiable via automatic differentiation and ASM to handle the solver's iterative loops \cite{DifferentiablePhysicsReview, Degrave2021}. The solver's output feeds into the loss function, allowing gradients to backpropagate through the solver to update network parameters.

\begin{figure}[!ht]   
        \centering
        \includegraphics[width=0.6\linewidth]{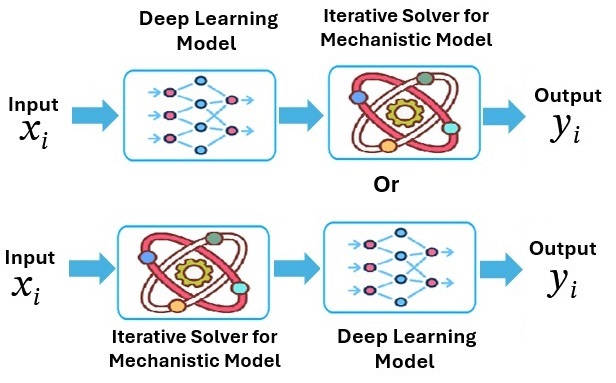}
     \caption{Series or Solver in the Loop Model-Architectural Configuration}
     \label{Series Model}
\end{figure}

Figure \ref{Series Model} illustrates an architecture that enables end to end learning, permitting the model to produce physically plausible inputs or deduce latent system attributes from empirical observations. In contrast to purely data-driven methods, incorporating a solver within the computational loop ensures adherence to foundational physical principles. For complex biological domains such as brain biomechanics, where precise and computationally tractable simulations are paramount integrating a differentiable solver into deep learning frameworks yields predictions that are both mechanistically accurate and data driven. This paradigm bears substantial potential for propelling neuro-biomechanical modeling via the fusion of mechanistic models and machine learning efficiency.\\
\\
The utility of a solver in the loop is particularly evident in:
\begin{itemize}
    \item \textbf{Inverse Problem Solving Physics Laws:} Inferring complex material properties or initial conditions directly from observed data, where the simulation model itself is part of the differentiable pipeline \cite{liang2019differentiable}. This is critical for patient specific neurological modeling, where sparse clinical measurements must inform complex biomechanical parameters.
    \item \textbf{Data Assimilation and State Estimation:} Integrating real time observations into physical models to refine state predictions, where the solver evolves the system and the learning model updates it based on incoming data.
     \item \textbf{Learning Controllers for Physical Systems:} Designing optimal control strategies by differentiating through the system's dynamics, allowing the deep learning model to learn control policies that directly influence a physical simulation \cite{Holl2020Learning, Qiao2020ScalableDiffPhysics, de2018end}.
\end{itemize}
\paragraph{Training Regimen}
The solver in the loop architecture, as discussed, integrates a classical numerical solver directly into the differentiable pipeline of a hybrid model. The training process for such configurations aims to optimize parameters that influence the physical system (e.g., material properties, boundary conditions, or even parts of the governing equations) by minimizing a defined loss function that compares the solver's output with observed data. This end to end differentiability empowers data driven learning within physically constrained environments.

The training process unfolds as follows:
\begin{enumerate}
    \item[TS-1] \textbf{Parameterization and Forward Inference}
    Training begins with an input, which might represent initial conditions, system parameters, or boundary conditions. A deep learning component (e.g., a neural network) may process this input to generate or refine parameters that directly feed into the classical numerical solver. For example, the neural network might predict the spatially varying elasticity modulus of brain tissue. The classical solver then executes its forward simulation using these parameters. It solves the underlying differential equations (e.g., a Finite Element Method simulation for brain deformation or a fluid dynamics solver for CSF flow) to produce the simulated physical outputs (e.g., displacement fields, stress distributions, fluid velocities). This step constitutes the core of the solver in the loop as the numerical simulation is explicitly run.

    \item[TS-2] \textbf{Loss Calculation}
    A loss function is defined to quantify the discrepancy between the simulated outputs generated by the classical solver and the corresponding empirical observations or target values. This could be a data fidelity term (e.g., mean squared error against measured displacements) or incorporate regularization terms.

    \item[TS-3] \textbf{Gradient based Optimization through the Solver}
    This is the critical step that enables training. To update the parameters of the deep learning model (and potentially the physical parameters fed into the solver if they are also trainable), the gradients of the loss function with respect to these parameters must be computed. Since the classical solver's operations are part of the computational graph, automatic differentiation techniques are applied to propagate gradients backward through the entire solver \cite{Degrave2021}. For complex, iterative solvers or simulations involving PDEs, this often relies on sophisticated methods like the ASM (as used in Neural ODEs but generalized to other solvers) or by carefully crafting custom differentiation rules for the solver's operations \cite{DifferentiablePhysicsReview}. This ensures that the computed gradients accurately reflect how changes in the input parameters influence the final simulation output. Finally, standard gradient based optimization algorithms (e.g., Adam, SGD) utilize these computed gradients to iteratively adjust the deep learning model's parameters (and any other trainable physical parameters). This process continues until the loss function is minimized, indicating that the model's predictions align well with the observed data while respecting the underlying physical laws enforced by the solver.
\end{enumerate}
The training of solver in the loop models is computationally intensive but offers significant advantages: it guarantees physical consistency of predictions, facilitates robust inverse problem solving, and can lead to more generalizable models than purely data driven approaches, especially vital for patient specific modeling and scientific discovery in biomechanics and neurological sciences where physical realism is paramount.
\paragraph{Advantages}
\begin{enumerate}
    \item[A1] \textbf{Gradient Flow Through Solvers} Unlike PINN, solver in the loop approach backpropagate gradients directly through solver, ensuring accurate and stable  parameter updates in the physically meaningful range.
    \item[A2] \textbf{Generalization With Less Data} Due to physical law enforcement by the mechanistic solver, the data requirement for training purposes is comparatively less in respect of purely data driven models. This makes the configuration highly valuable for domains in which obtaining huge labeled data is not cost effective or impractical.
    \item[A3] \textbf{Reduce Numerical Errors} To reduce computational cost and accelerate the solver, it is preferred to utilize low fidelity mechanistic model. However; a low fidelity mechanistic model can incur large numerical errors. In solver in the loop configuration, the neural network can be trained with high fidelity solution as labeled data to transform the low fidelity outcomes and reduce numerical errors.
 \begin{figure}[!ht]
    \centering
    \captionsetup{justification=centering,margin=2cm}
    \includegraphics[width=\linewidth, trim={0cm 5.2cm 0cm 6.5cm}, clip]{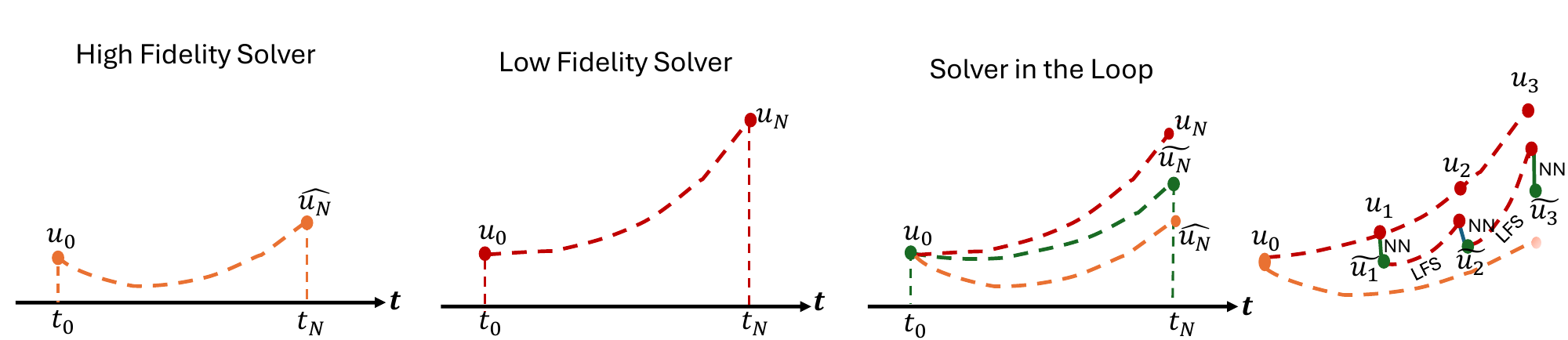}
    \caption{Transformation of low fidelity solution trajectory (red) towards high\\ fidelity solution trajectory (orange) reducing numerical errors \cite{Um2020Solver}}
    \label{Reduce Numerical Errors}
\end{figure}
\end{enumerate}
\paragraph{Challenges and Adaptation Strategies}
In hybrid modeling, another series configuration directly embeds numerical solvers inside the training loop. Unlike NODEs, this setup directly backpropagates through physics solvers (e.g., PDEs, rigid body dynamics, or molecular dynamics). Here, the neural network provides parameters, forcing terms, or closure relations for a physics based solver, and gradients are propagated through solver iterations \cite{Chen2025, Tathawadekar2023}. This setup is particularly attractive for scientific domains such as biomechanics, fluid dynamics, and multiphysics forcasting. While being highly expressive and ensuring physically faithful training, this configuration introduces severe computational and implementation challenges as follows:  

\begin{enumerate}
    \item[C1] \textbf{Solver instability and stiffness} Embedding neural parameters into nonlinear PDE or ODE solvers often increases stiffness, leading to solver divergence or excessive refinement steps \cite{BRADLEY2022}. Adaptive solvers may require many function evaluations, especially in stiff systems, leading to high computational costs \cite{Kidger2020NeuralCDEs}.\\
    \textbf{Adaptation Strategies}\\
   Solver stiffness can be reduced by employing  regularization. To reduce the computational cost incurred due to excessive refinement steps, one should utilize a surrogate assisted training (pretraining on coarse solvers before refinement) \cite{bhatia2025prdpprogressivelyrefineddifferentiable}, implicit differentiation to bypass full unrolling of solver iterations \cite{LeCleach2022Dojo, schnell2024stabilizingbackpropagationtimelearn, LIST2025}, and solver aware regularization to ensure well conditioned dynamics \cite{Chen2025}.
    \item[C2] \textbf{Gradient quality} Differentiating through solvers can cause exploding or vanishing gradients, especially for long time horizons. Iterative solvers may yield unstable or noisy gradients, especially for implicit schemes or contact dynamics \cite{Hu2019DiffTaichi}.\\
    \textbf{Adaptation Strategies}\\
    Exploding or vanishing gradients can be caused due to long temporal dependency, unsuitable parameter initialization, saturating activation function. One can avoid such problem by using suitable non saturating activation function, clipping or scaling technique, appropriate parameter initialization and modifying the neural architecture.  
    \item[C3] \textbf{Hyperparameter explosion and Computational scalability} In addition to the design of neural architecture, solver tolerances, iteration limits, and discretization strategies become critical hyperparameters \cite{SCHWEIDTMANN2024}. Solver in the loop models are significantly slower than residual models due to repeated high fidelity simulations during training. Standard machine learning frameworks are not optimized for differentiable solvers, making efficient implementations non trivial.\\ 
    \textbf{Adaptation Strategies}\\
    To handle a huge number of hyperparameters at a time and for efficient implementation of solver in the loop architecture, domain specific differentiable physics libraries have been developed in recent years \cite{Werling2018NimblePhysics, Hu2019DiffTaichi, Freeman2021Brax, Krishna2021GradSim, LeCleach2022Dojo, Holl2020PhiFlow, Schoenholz2020JAXMD, Xian2023FluidLab, Lin2022daXBench, Makoviychuk2021IsaacGym, Doerr2022TorchMD, Thoelke2022TorchMDNet, Pelaez2024TorchMDNet2.0, PhysicsNeMo2025, ren2023diffmimicefficientmotionmimicking, innes2019differentiable, rackauckas2019diffeqfluxjljulialibrary}. 
    
    Neurological disorders modeling is linked  with different scientific domains such as fluid dynamic, molecular dynamic, material science, contact dynamic. In most cases of neurological disorders, gait cycle or human body movement is affected. This impairment can provide an insight into severity of disorder and rehabilitation requires artificial support. In interest of incorporating simulations from diverse scientific domains in differentiable framework, domain specific differentiable physics libraries have been primarily developed. For the computation scalability, the efficient adjoint method and GPU based acceleration is supported by the libraries such as DiffTaichi \cite{Hu2019DiffTaichi}, $\Phi$Flow \cite{Holl2020PhiFlow}, and JAX MD \cite{Schoenholz2020JAXMD}. NimblePhysics \cite{Werling2018NimblePhysics} provides efficient way of gradient calculation by incorporating Lagrangian dynamics for making rigid body dynamics with contact differentiable. For molecular dynamics, Jax MD \cite{Schoenholz2020JAXMD}, torchmd \cite{Doerr2022TorchMD}, TorchmdNet \cite{Thoelke2022TorchMDNet}, TorchmdNet 2.0 \cite{Pelaez2024TorchMDNet2.0} can be utilised for simulation acceleration and neural network integration for different tasks. GradSim ($\nabla$Sim) \cite{Krishna2021GradSim} can be employed for differentiable rendering with multiphysics simulation which can utilise visual data directly for training. For hemodynamical or other fluid dynamic simulations, $\Phi$Flow \cite{Holl2020PhiFlow}, Fluidlab \cite{Xian2023FluidLab}, DiffTaichi \cite{Hu2019DiffTaichi} have been developed. For human motion analysis, Brax\cite{Freeman2021Brax}, Dojo \cite{LeCleach2022Dojo}, Isaac Gym \cite{Makoviychuk2021IsaacGym} have been created. Major shift has been witnessed in integrating simulations into machine learning framework due to development of such dedicated libraries.
\end{enumerate}
Both \textbf{Neural ODEs} and \textbf{solver in the loop frameworks} contribute substantially to Hybrid modeling, but they serve different roles. Neural ODEs are most effective in reducing memory use, offering adaptable accuracy efficiency trade offs, and modeling continuous time processes with irregular data. By contrast, solver in the loop methods emphasize adherence to physical laws, improving robustness and generalization, particularly when data is limited. In short, Neural ODEs prioritize \textbf{computational flexibility}, whereas solver in the loop strategies focus on \textbf{physical fidelity and generalization}. Together, they provide complementary pathways for integrating machine learning with physical modeling.
\subsection{Series-Parallel Hybrid Models}
\begin{figure}[!ht]
    \centering
    \includegraphics[width=\linewidth]{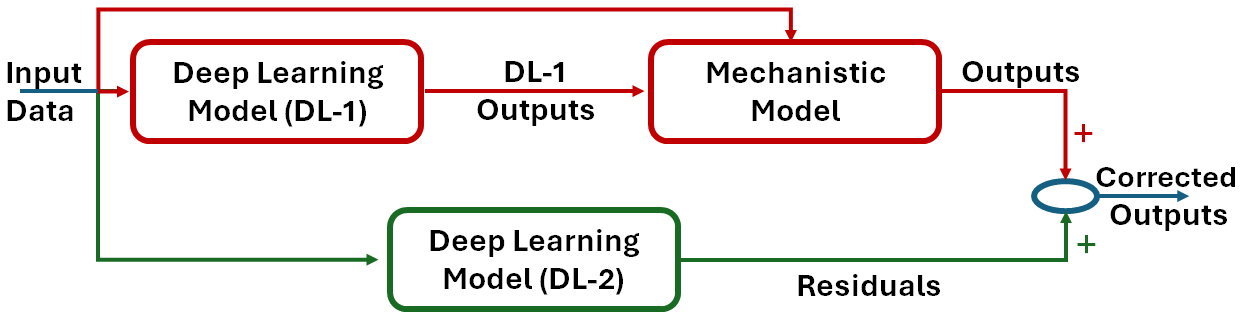}
    \caption{Parallel-Series Architecture, the parallel path (Green), the series path (Red) \cite{Bhutani2006}.}
    \label{Parallel-Series Architecture}
\end{figure}

The parallel–series architecture takes the strengths of the standalone parallel and series models together to complement each other as delineated in Figure \ref{Parallel-Series Architecture}. The series path (red) provides flexibility to the mechanistic model utilizing the deep learning model in series to process the raw input data and make it suitable for the mechanistic model by dimensional reduction and nonlinear transformation. In parallel path (green), the second deep learning model is augmented to learn residual of the mechanistic model. This residual model acts as a correction mechanism, capturing unmodeled nonlinearities, stochastic disturbances, or systematic biases that cannot be explained by physics alone. Together, both paths guarantee flexibility, stability, certainty, and synchronization. The model offers a synergistic coupling of the deep learning’s predictive power with the structural consistency of the mechanistic principals. This layered configuration is deliberately modular, making it easier to calibrate individual components, enforce physics based regularization in the training loop, and scale the approach to large, high dimensional dynamical systems where both interpretability and adaptability are essential.\par

\subsubsection{Training Regimen}
The training of parallel-series configuration is aimed at optimizing learnable parameters of two deep learning models, one responsible for residual correction and the other for calibration and acceleration of the mechanistic model. As can be observed in Figure , the flow of forward calculation flows through two parallel paths , and therefore the gradient will also flow back along these two paths for optimization. The training procedure unfolds in three training steps as follows.
\begin{enumerate}
    \item[TS-1] \textbf{Mechanistic Calibration} The training process begins by estimating or fitting the parameters of mechanistic model using well established historical data to ensure physical consistency. Once the model parameters are chosen, the first deep learning model DL-1 is trained to transform the raw input data into effective model inputs that improve compatibility with the first principles solver for accuracy and acceleration. Training is carried out by optimizing the physical loss between the mechanistic model outputs and observed data or higher fidelity outputs.
    \item[TS-2] \textbf{Residual Learning} To capture unmodeled physics, stochastic noise, and data bias, the second deep learning model DL-2 is trained from residual loss. This training will reduce the mismatch between prediction and observation without coupling with a mechanistic solver.  
    \item[TS-3] \textbf{Joint Optimization} For end to end training of the parallel-series model, the composite loss must be generated that accounted for prediction accuracy, residual control, and compliance with the conservation law together. By optimization of this composite loss, both previously trained deep learning models will be fine tuned for the whole architectural synchronization. 
\end{enumerate}
\subsubsection{Advantages}
The model offers synergic coupling of the deep learning's predictive power with the structural consistency of mechanistic principals.
\begin{enumerate}
    \item[A1] \textbf{Bias Correction and Robustness}
    Unknown physics and nonlinear secondary interactions of complex dynamical systems can be captured by residual model while keeping the consistency of governing laws intact through series model. This way, the model delivers better predictive accuracy by reducing structural bias under partial observability. The model also provide real time access to latent variables through auxiliary deep learning model which maps measurable states to unmeasurable states \cite{KANEKO2022}. This makes the model perfect choice for real life applications where observable data is not only sparse but also noisy introducing considerable uncertainty \cite{Liu2021}.
    \item[A2] \textbf{Scalability}
    In engineering and healthcare domains, both accuracy and interpretability plays a crucial role for reliable adoption. The parallel series model provides accuracy due to residual correction component and interpretability through series model, hence makes it perfectly scalable across diverse application domains.
\end{enumerate}
Taken together, these advantages make the parallel–series model a promising tool in the pursuit of reliable, adaptable, and scalable modeling for complex dynamical systems; however, the implementation of parallel-series model is challenged by the complexity of the model architecture, stability, and computational efficiency.

\subsubsection{Challenges and Adaptation Strategies}
Parallel–series models combine residual learning with solver driven integration, aiming to capture unmodeled dynamics while maintaining physically consistent trajectories \cite{Chen2025, Tathawadekar2023}. Although conceptually appealing, this architecture inherits compounded challenges:
\begin{enumerate}

\item[C1] \textbf{Model complexity} Integrating residual learners with series models increases the number of hyperparameters, such as solver tolerances, architecture depth, and regularization weights \cite{SCHWEIDTMANN2024}. Such models are difficult to train and interpret.  
\item[C2] \textbf{Generalization}  Parallel–series models excel in interpolative tasks and perform well within known regimes; however, under unseen operating conditions, the extrapolation by the model may not be reliable unless strong physical priors are explicitly enforced \cite{KANEKO2022}. The numerical errors might be amplified due to feedback between the physics model and the learned components, which can result in destabilization of the series model by residual corrections \cite{BRADLEY2022}.
\end{enumerate} 
\textbf{Adaptation Strategies}\\ 
To overcome the hyperparameter explosion, modular training strategies can be utilized in which residual and solver based components are calibrated independently before integration. To stabilize the series model and improve generalization, the composite loss of parallel-series model should be jointly optimized under strong physics constraints, which empower the penalization for violation of conservation laws.

Taken together, the obstacles that hybrid modeling  faces not only arises from solver inefficiency or training instability but also from the need to balance physical fidelity, neural flexibility, and computational feasibility. Parallel models are efficient but susceptible to bias; series models are elegant but computationally expensive; and parallel-series models are powerful but complex to tune. The emerging consensus is that no single configuration suffices universally. Instead, the future lies in adaptive strategies that integrate neural approximators with first principles solvers in a solver aware, domain informed manner \cite{BRADLEY2022,SCHWEIDTMANN2024}. This perspective underscores the importance of hybridization as a flexible toolkit, where the interplay between physics and deep learning is guided by problem context rather than dictated by a single modeling philosophy.
\section{Making Mechanistic Tumor Growth Model Differentiable}
In the context of hybrid modeling of Neurological disorders, the objective is to bridge the mechanistic model with the deep learning framework, either in parallel or series configuration. For this, we need to make the mechanistic simulation differentiable so that the gradient information can flow through both the model and the deep learning components. To make the mechanistic simulation differentiable, we must preserve the solution procedure as a computational graph as delineated in Figure  for the toy problem.  We need to either implement every solver operation with autodiff capable primitives so that an autodiff engine can backpropagate through the entire sequence of operations or leverage the adjoint state method to get the gradient information by solving a separate backward in time differential equation governing the adjoint state without explicitly backpropagating through the full computational graph of the forward dynamics. It is required to build a custom backward gradient rule using the Adjoint State method.
Here, an attempt has been made to explain the procedure through the following toy problem.
\subsection{Toy Problem} Let us consider tumor growth modeled using a reaction–diffusion partial differential equation as discussed by Kapteyn et al.\cite{kapteyn2025tumortwinpythonframeworkpatientspecific} which describes the temporal evolution of tumor cellularity, driven by cell invasion (diffusion), logistic proliferation (reaction), and treatment effects (radiotherapy and chemotherapy).
\begin{align}
    \frac{\partial u(x,t)}{\partial t} =& \nabla\cdot(\kappa \nabla u(x,t)) + \gamma(x) u(x,t) \left(1 - \frac{u(x,t)}{K}\right)\nonumber\\ &-  \sum_{p=1}^{n_{CT}}\sum_{q=1}^{T_p} \eta_p \chi_p \exp(-\zeta_p(t - \tau_{p,q})) u(x,t)
\end{align}
\begin{align}
    u(x,t)_{\text{after}}=u(x,t)_{\text{before}}\exp\left( -\rho d(t) - \omega d^2(t) \right)
\end{align}
Here, $u$ denotes the tumor cellularity, $\kappa$ is the diffusion coefficient, $\gamma(x)$ is the net proliferation rate. The parameter $K$ denotes the carrying capacity and $n_{CT}$ is the Number of different chemotherapy agents. Furthermore, $T_p$ represents the total number of doses administered for agent $p$, while $\eta_p, \zeta_p$ denote the sensitivity and decay parameters for the  chemotherapy agent $p$ respectively. The parameter $\chi_p$ represents the efficacy of the chemotherapy agent $p$. $\tau_{p,q}$ denotes the timing of $q^{th}$ dose for the $p^{th}$ agent. As radiotherapy can be modeled as an instantaneous treatment implying an immediate change in tumor cellularity. This effect is represented by $u(x,t)_{\text{before}}$ and $u(x,t)_{\text{after}}$, which denote the tumor cellularity immediately before and after the administration of a radiotherapy dose $d(t)$, governed by the radio sensitivity parameters $\rho$ and $\omega$.
Under the assumption of $\kappa$ and $\gamma(x)$ being homogeneous throughout the spatial domain, and combining radiotherapy response into continuous dynamics denoting the instantaneous treatment time as $t_{RT}$, One can rewrite equations (6)-(7), as 
\begin{align}
    \frac{\partial u(x,t)}{\partial t} =& \kappa \nabla^2 u(x,t) + \gamma u(x,t) \left(1 - \frac{u(x,t)}{K}\right)\nonumber\\ &-  \sum_{p=1}^{n_{CT}}\sum_{q=1}^{T_p} \eta_p \chi_p \exp(-\zeta_p(t - \tau_{p,q})) u(x,t) \nonumber\\ &- \delta(t - t_{RT}) \left( \rho d + \omega d^2 \right) u(x,t)
\end{align}
Where $\delta(t - t_{RT})$ is a Dirac delta function and model an instantaneous radiotherapy treatment at time $t=t_{RT}$.\\
\subsubsection{Spatial Discretization}
Equation (8) can be spatially discretized using a second order central difference scheme with grid size corresponding to the voxel size of input MRI, the continuous spatial derivatives are replaced by differences between neighboring voxels.

\begin{align}
\frac{d u_{i,j,k}(t)}{dt} = \kappa \Biggl[ &\frac{u_{i+1,j,k}(t) - 2u_{i,j,k}(t) + u_{i-1,j,k}(t)}{\Delta x^2} + \frac{u_{i,j+1,k}(t) - 2u_{i,j,k}(t) + u_{i,j-1,k}(t)}{\Delta y^2}\nonumber\\ &+ \frac{u_{i,j,k+1}(t) - 2u_{i,j,k}(t) + u_{i,j,k-1}(t)}{\Delta z^2} \Biggr] + \gamma u_{i,j,k}(t) \left(1 - \frac{u_{i,j,k}(t)}{K}\right) \nonumber\\&- \left( \sum_{p=1}^{n_{CT}}\sum_{q=1}^{T_p} \eta_p \chi_p \exp(-\zeta_p(t - \tau_{p,q})) \right) u_{i,j,k}(t)\nonumber\\ &- \delta(t - t_{RT}) \left( \rho d + \omega d^2 \right) u_{i,j,k}(t)
\end{align}
\subsubsection{Time Discretization}
Equation (9) can be further discretized using a forward Euler scheme as
\begin{align}
u_{i,j,k}^{m+1} = u_{i,j,k}^{m} + \Delta t \Biggl[ 
&\kappa \left( \frac{u_{i+1,j,k}^{m} - 2u_{i,j,k}^{m} + u_{i-1,j,k}^{m}}{\Delta x^2} + \frac{u_{i,j+1,k}^{m} - 2u_{i,j,k}^{m} + u_{i,j-1,k}^{m}}{\Delta y^2}\right. \nonumber\\
& \left. + \frac{u_{i,j,k+1}^{m} - 2u_{i,j,k}^{m} + u_{i,j,k-1}^{m}}{\Delta z^2} \right) + \gamma u_{i,j,k}^{m} \left(1 - \frac{u_{i,j,k}^{m}}{K}\right) \nonumber\\ &- \left( \sum_{p=1}^{n_{CT}}\sum_{q=1}^{T_p} \eta_p \chi_p \exp(-\zeta_p(t_m - \tau_{p,q})) \right) u_{i,j,k}^{m} - \delta(t_m - t_{RT}) \left( \rho d + \omega d^2 \right) u_{i,j,k}^{m}
\Biggr]
\end{align}
Here, to ensure the numerical stability and the instantaneous time point of radiotherapy to be captured as a discrete time step,  $\Delta t$ must be chosen which satisfies $\Delta t \le \frac{1}{2\kappa \left( \frac{1}{\Delta x^2} + \frac{1}{\Delta y^2} + \frac{1}{\Delta z^2} \right)}$ and $t_{RT} \pmod{\Delta t} = 0$.
The model parameters are $P=\{\kappa, \gamma, K, \eta, \zeta, \rho, \omega\}$, where $\eta=[\eta_1,\cdots,\eta_{n_{CT}}]$ and $\zeta=[\zeta_1,\cdots,\zeta_{n_{CT}}]$. To initiate the model, initial cellularity $u^{0}$ in each voxel can be derived by the apparent diffusion coefficient (ADC) of the diffusion weighted MRI of the patient at the first visit and model parameters will be initialized with initial guess. 
The model will run from $t=0$ to $t=M$, 0 represents the first visit of the patient, and $M$ represents the second visit of patient. As depicted in Figure, at the completion of forward pass through the model, cellularity of Tumor region will be predicted and at the same time point cellularity of Tumor region can be observed through the the ADC of the diffusion weighted MRI of the patient at the second visit. From observed and predicted voxel wise cellularity, Total Tumor Cell count (TTC) can be obtained and by comparing them the loss $\mathcal{L}$ will be generated.\\
The gradient of loss $\mathcal{L}$ with respect to model parameters $P$ can be used for optimizing parameters in gradient based optimization scheme. One way to obtain such gradient is to use a solver developed in a differentiable programming environment to solve equation (10) and use auto differentiation for backpropagation of the gradient. However, it is not a feasible choice in respect of memory as for backpropagation entire forward trajectory need to be stored. The alternative is to use adjoint state method (ASM), which has been utilized by Kapteyn et al.\cite{kapteyn2025tumortwinpythonframeworkpatientspecific}  with DiffTaichi. Here an effort has been made to explain the entire working flow of ASM as follow. 

\section{Prospects in Neurological Disorder Modeling}
Hybrid modeling has recently emerged as a powerful methodological paradigm for studying neurological disorders, providing a principled means of integrating biophysical knowledge with modern machine learning techniques. By embedding differential equation–based descriptions within trainable models, this paradigm enables disease processes to be represented as continuous time dynamical systems. Unlike purely data driven, discrete time approaches, differentiable models inherently respect the temporal continuity and causal structure of biological phenomena. Among these approaches, Neural Ordinary Differential Equations (Neural ODEs)\cite{Chen2018NeuralODE} and Universal Differential Equations (UDEs) \cite{DifferentiablePhysicsReview}  have attracted increasing attention due to their ability to infer latent disease dynamics directly from longitudinal data while maintaining interpretability through explicit dynamical formulations \cite{Cai2023, Han2024, li2025exploringneuralordinarydifferential, ElGazzar2024, PHILIPPS202425}.

A key advantage of hybrid modeling lies in its capacity to unify prediction and system identification. Such models can extrapolate future disease trajectories while simultaneously estimating unobserved initial states or physiological parameters governing disease evolution. Recent studies have demonstrated the applicability of Neural ODEs in modeling tumor growth dynamics and capturing  spatio-temporal progression patterns in neurodegenerative disorders such as geographic atrophy and Alzheimer’s disease \cite{Jaroudi2023,Lachinov2023}. Complementarily, UDEs have been proposed as a unifying modeling language for neuroscience, enabling the integration of mechanistic equations derived from known biological principles with data driven components that account for unknown or poorly characterized processes \cite{ElGazzar2024}. Together, these approaches provide a flexible yet principled framework for studying complex brain disorders.

More broadly, hybrid modeling offers a synthesis of predictive accuracy and mechanistic interpretability. By incorporating prior biological knowledge, accommodating sparse and irregular clinical observations, and representing disease evolution as a continuous dynamical process, these models hold substantial promise for patient specific forecasting, treatment optimization, and improved mechanistic understanding of neurological disease progression.
\subsection{Brain Tumor Progression and Neurological Outcome Prediction}
Patients with brain tumors frequently experience neurological impairments, including cognitive decline, gait disturbances, and seizures. Predicting tumor progression and optimizing therapeutic strategies remain challenging due to the infiltrative growth patterns and marked heterogeneity of these malignancies. Consequently, reliable mathematical models are essential for accurate prognosis and clinical decision making. A common starting point involves ordinary differential equation (ODE) models that describe the temporal evolution of tumor burden or cellularity,
\begin{align}
\frac{dN}{dt} &= f(N(t), u, t), \nonumber\\
N(0) &= N_{\text{initial}},
\end{align}
where $N(t)$ denotes a time dependent tumor related state variable, $u$ represents treatment inputs, and $f(\cdot)$ captures growth and treatment effects. A typical formulation decomposes tumor dynamics into proliferation and treatment induced reduction,
\begin{align}
f(N(t),u,t) &= P(N,u_t,t)N(t) - \mu(t,u_t)N(t), \quad t \notin t_{\text{treatment}}, \nonumber\\
N(t_{\text{treatment}}^+) &= S(u_{t_{\text{treatment}}})N(t_{\text{treatment}}^-), \quad t \in t_{\text{treatment}},
\end{align}
where $P(\cdot)$ models proliferation, $\mu(\cdot)$ represents gradual treatment effects, and $S(\cdot)\in[0,1]$ denotes the survival fraction associated with discrete treatment events.

ODE based tumor models are computationally efficient and interpretable and have been successfully used in personalized digital twin frameworks. For instance, Chaudhuri et al.\cite{ChaudhuriPredictivedigitaltwin} employed Bayesian calibration of an ODE model using MRI derived tumor burden measurements to enable uncertainty aware forecasting and adaptive radiotherapy planning. However, purely temporal models are unable to capture spatial heterogeneity or tumor infiltration across distinct brain regions, which are critical determinants of neurological outcome. As a result, ODEs alone are insufficient for comprehensive tumor modeling.

To address these limitations, partial differential equation (PDE) formulations are employed to capture  spatio-temporal tumor dynamics, including invasion, proliferation, and treatment response,
\begin{align}
\frac{\partial N(x,t)}{\partial t}
= -\nabla \cdot (-D\nabla N(x,t))
+ P(N(x,t),t)
- T(N(x,t),t).
\end{align}
where $N(x,t)$ denotes a spatio-temporal dependent tumor related state variable, $D$ represents diffusivity constant i.e. the invasive spreading in neighboring healthy region, $P(\cdot)$ models proliferation, and $T(\cdot)$ represents treatment response. Extensions of this framework incorporate biomechanical coupling to account for tissue deformation and stress mediated growth inhibition.

\begin{figure}[!ht]
    \centering
    \includegraphics[width=0.9\linewidth]{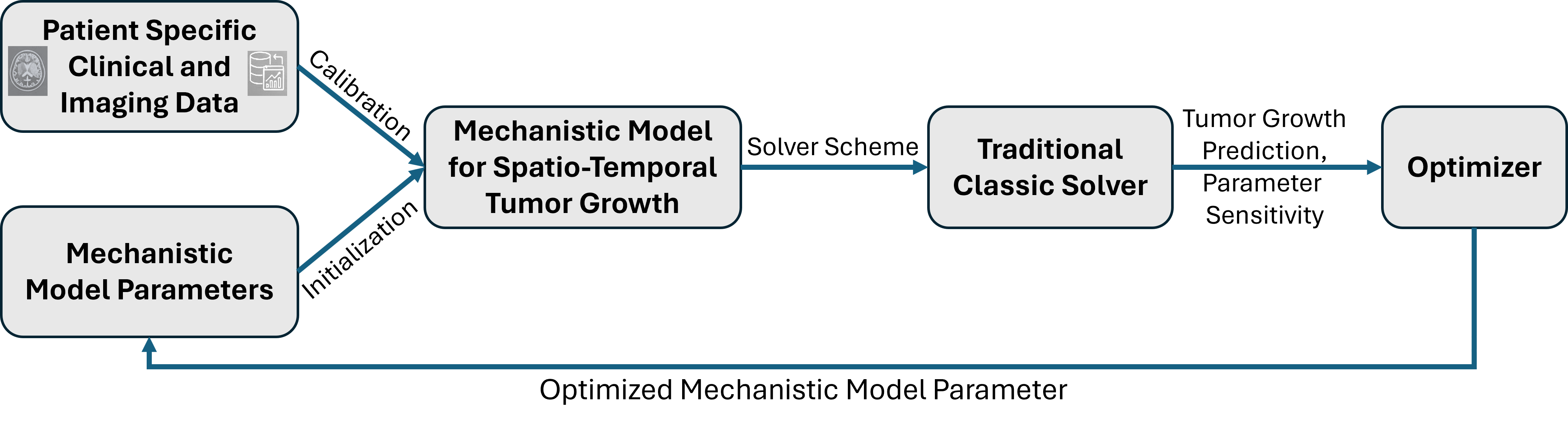}
    \caption{TumorTwin key components \cite{kapteyn2025tumortwinpythonframeworkpatientspecific}.}
    \label{fig:tumortwin}
\end{figure}

Kapteyn et al. \cite{kapteyn2025tumortwinpythonframeworkpatientspecific} introduced \textit{TumorTwin}, a modular Python framework for patient specific digital twins based on reaction diffusion PDEs initialized from imaging and treatment history. While the framework enables individualized forecasting, its reliance on simplified growth laws and coarse tumor burden measurements limits its ability to fully exploit high resolution spatial imaging data.

Data driven continuous time models offer a complementary solution. Xiang et al.\cite{Xiang2024} proposed a latent Neural ODE framework in which patient covariates are encoded into probabilistic latent initial conditions and evolved through an ODE solver to predict tumor dynamics and progression free survival. Extending this paradigm to neuro oncology, hybrid Neural ODE models can integrate multimodal inputs—such as MRI derived features, treatment schedules, and neurocognitive assessments—to generate personalized forecasts of tumor progression and neurological decline. Similarly, Mohammed and Mohanty \cite{Mohammed2025} developed a hybrid model combining deterministic, stochastic, and delay dynamics with Neural ODEs to accurately capture tumor drug interactions, highlighting the potential of these methods for adaptive trial design and individualized therapy planning.

More recently, Laslo et al.\cite{laslo2025mechanisticlearningguideddiffusion} proposed a mechanistic learning framework that couples an ODE based tumor growth model with a gradient guided denoising diffusion implicit model to generate spatio-temporal MRI forecasts. By conditioning a generative diffusion process on mechanistic priors, the approach produces anatomically consistent predictions while reducing dependence on large longitudinal datasets.

\subsection{Neurodegenerative Disease Progression}

Neurodegenerative disorders exhibit structured patterns of pathological spread and region specific tissue loss, suggesting that disease evolution is governed by underlying biological and physical principles rather than purely statistical associations \cite{Weickenmeier2018Multiphysics, Weickenmeier2019PhysicsBased, DaSilva2021BiomechanicalDeep}. Mechanistic models capture these processes by linking molecular pathology to macroscopic brain changes through systems of differential equations. At a coarse scale, compartmental ODE models describe interactions among interconnected brain regions,
\begin{equation}
\frac{d x_i(t)}{dt} = f_i(x_i(t)) + \sum_j g_{ij}(x_j(t),x_i(t)),
\end{equation}
where $x_i(t)$ denotes a regional pathological or structural variable of $i^{th}$ brain compartment, $f_i$ denotes local disease dynamics within the $i^{th}$ brain compartment and  $g_{i,j}$ represents the interaction between $i^{th}$ and $j^{th}$ brain compartment. To resolve spatial heterogeneity, reaction diffusion PDE models are employed to describe the propagation of toxic or misfolded proteins,
\begin{equation}
\frac{\partial c(\mathbf{x},t)}{\partial t}
= \nabla \cdot \left( \mathbf{D}(\mathbf{x}) \nabla c(\mathbf{x},t) \right)
+ R\big(c(\mathbf{x},t)\big),
\end{equation}
Here, $\mathbf{x}$ represents spatial position in the brain, $c(\mathbf{x},t)$ denotes concentration of misfolded or toxic protein, $D$ is the diffusion constant and $R(\cdot)$ denotes the local reaction (production, clearance, aggregation). Such model are often coupled with biomechanical formulations linking pathology to tissue atrophy,
\begin{equation}
\nabla \cdot \boldsymbol{\sigma}(\mathbf{u}) + \mathbf{b}(c(\mathbf{x},t)) = \mathbf{0}.
\end{equation}
 where, $\mathbf{u}$ denotes tissue displacement field, $\boldsymbol{\sigma}(\mathbf{u})$ is stress tensor due to deformation, and $\mathbf{b}(\cdot)$ is the pathology induced body force. While these models provide mechanistic interpretability, their clinical applicability is constrained by computational cost, strong modeling assumptions, and difficulties in parameter personalization \cite{Khanal2016Biophysical, Blinkouskaya2022Thesis}. These limitations have motivated the integration of deep learning with continuous time modeling.

Lachinov et al.\cite{Lachinov2023} introduced a Neural ODE framework for predicting spatio-temporal disease progression from longitudinal imaging, using a single baseline scan to forecast future anatomical changes. Similarly, Wen \cite{Wen2020} proposed a Neural ODE–based framework for modeling resting state fMRI dynamics, enabling both forward prediction and temporal interpolation while providing a principled and interpretable description of latent brain dynamics. Das et al. \cite{DAS2025} recently propose a purely data driven, interpretable Bayesian Encoder–Decoder GRU (BEND-GRU) model to predict Alzheimer’s disease progression from longitudinal clinical data, focusing on uncertainty quantification and temporal prediction . By integrating a Bayesian approach with a recurrent neural network architecture, the model captures temporal patterns in patient data and provides interpretable estimates of how confident it is in its predictions. This probabilistic framework aims to enhance clinical reliability compared with traditional deep learning models by explicitly modeling uncertainty, which is especially important in progressive diseases like Alzheimer’s where data can be noisy and heterogeneous. The authors’ work highlights an opportunity for hybrid modeling, where data driven Bayesian learning is combined with physics based constraints to achieve both predictive accuracy and biologically meaningful interpretation of Alzheimer’s disease progression \cite{DAS2025}. 

As a purely data driven approach, the model is flexible and effective at learning patient specific patterns from heterogeneous data, but it does not explicitly model the physics  or biology based spatio-temporal propagation of pathology, as done in mechanistic models of amyloid or tau spread. In comparison, physics based models offer causal and spatial interpretability grounded in neuroscience but rely on strong assumptions and detailed biomarker data. A compelling future direction is hybrid modeling as delineated in Figure, which integrates data driven deep learning with physics based constraints, mechanistic models can enforce biologically plausible  spatio-temporal dynamics, while deep learning can adapt to patient specific variability, noisy observations, and incomplete data. Inspired by the architectures of \cite{Lachinov2023} and \cite{laslo2025mechanisticlearningguideddiffusion}, an attempt has been made to construct the serial + parallel architecture to not only predict the progression of Neuro degenerative disease but also generating the future MRI. In this model the initial data driven component will provide the disease severity state which will be utilized in mechanistic model with $\tau$ or $\beta$-amyloid pathology from genomics and other model parameters derived from patient MRI. Mechanistic model will provide the atrophy index to train the regressor applied over time embedded MRI stack. The scaling factor derived from regressor and mechanistic model output will predict the future atrophy map, which can be used with guided denoising diffusion model to generate future MRI. Such hybrid approaches have the potential to combine predictive performance, mechanistic insight, and uncertainty awareness, offering a more robust and clinically relevant framework for modeling the progression of Alzheimer’s disease.
\subsection{Incomplete Cerebrovascular Physiology}
    \begin{figure}[!ht]
        \centering
        \includegraphics[width=0.8\linewidth]{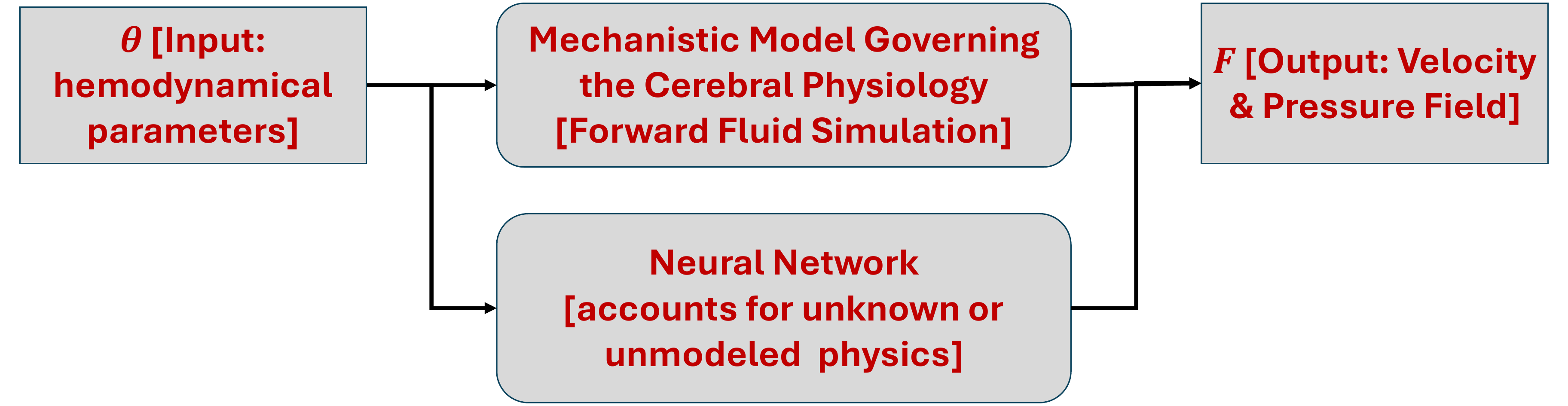}
        \caption{Hybrid Modeling for Missing or Incomplete Cerebrovascular Hemodynamical System}
        \label{Neural ODE for Cardiovascular System}
    \end{figure}
To understand neurological disorders such as stroke, dementia and traumatic brain injury, it is essential to study the cerebral hemodynamics, intracranial pressure regulation and blood brain interaction. To study such concepts, well established mathematical models of cerebral physiology can be employed. These models capture the core processes in simplified manner and as a result many processes like neurovascular coupling, metabolic regulation cerebrospinal fluid circulation and edema dynamics yet not properly and completely modeled. Due to such incomplete modeling, prediction accuracy can not be achieved and hinder the effective deployment in clinical setup. Recent advances in hybrid modeling offer a promising path forward. Grigorian et al. \cite{Grigorian2024} demonstrated a hybrid neural ODE model for the cardiovascular system.  A mechanistic ODE backbone was augmented with a neural network to capture ventricular interactions, and symbolic regression was employed to recover interpretable analytic expressions of the learned dynamics. In parallel, Demirkaya et al. \cite{Demirkaya2024} proposed a general hybrid ODE–neural network (NN) framework to model incomplete physiological systems, enabling joint estimation of hidden states, unknown parameters, and initial conditions while preserving mechanistic interpretability. The Hodgkin-Huxley neuron model  and the retinal circulation model were chosen as case studies to understand different feasible application scenarios. Inspired by these approaches, we propose a hybrid  framework for modeling missing or incomplete cerebrovascular hemodynamical system as illustrated in Figure, which is essential for accurately modeling neurological conditions such as brain strokes. Established ODEs represent well characterized vascular and cerebrospinal fluid mechanics, while neural networks approximate uncertain or highly nonlinear components using synthetic simulations and experimental or clinical data. Symbolic regression can be applied to transform learned dynamics into parsimonious analytic forms, thereby bridging data driven flexibility with mechanistic transparency. By systematically integrating first principles knowledge with neural components, this methodology addresses the challenge of incomplete physiology and paves the way for robust, interpretable, and patient specific modeling of cerebrovascular disorders.
\subsection{Universal Differential Equations (UDEs) for modeling in neuroscience}
 \begin{figure}[!ht]
        \centering
        \includegraphics[width=\linewidth]{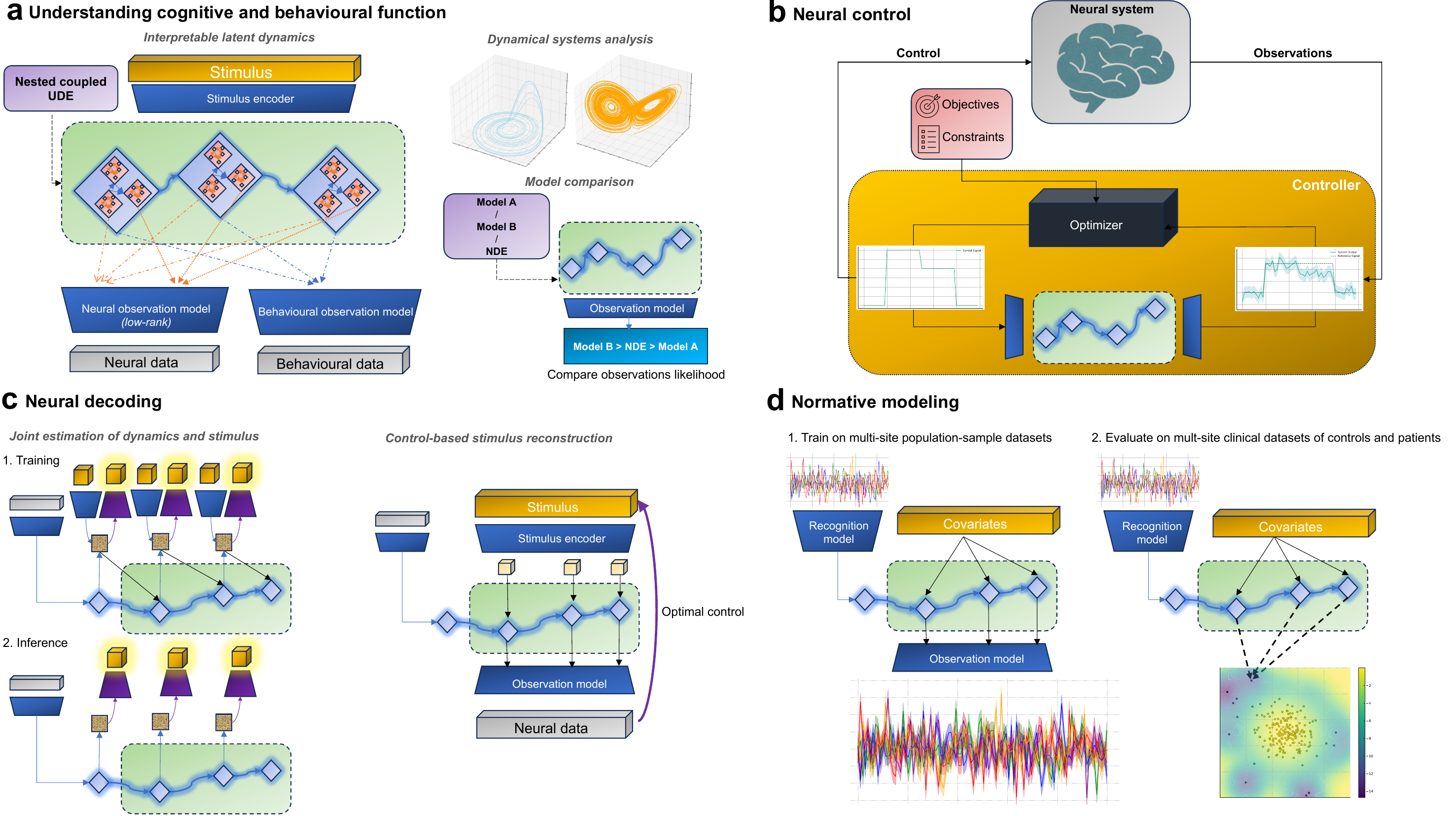}
        \caption{Universal Differential Equations as a
Common Modeling Language for Neuroscience (a)Uncover latent dynamics underlying cognition and behavior, (b) Closed loop neural control, (c) Neural decoding to reconstruct stimuli from neural recordings, (d) Normative modeling to model population level variability for patient stratification \cite{ElGazzar2024}}
        \label{UDE for Neuroscience}
    \end{figure}
ElGazzar and van Gerven \cite{ElGazzar2024} propose Universal Differential Equations (UDEs) as a powerful unifying framework for modeling in neuroscience. UDEs treat differential equations as flexible, differentiable structures that can be enriched with deep learning components, allowing them to combine the interpretability of mechanistic models with the adaptability of data driven methods. This synthesis bridges a long standing gap between mechanistic, phenomenological, and purely statistical approaches, creating a principled modeling language that can capture the complexity of neural systems. By embedding neural networks within classical dynamical systems, UDEs provide the flexibility to represent latent neural dynamics while retaining the structure needed to ensure biological plausibility and interpretability. This approach not only integrates decades of work in calculus, numerical analysis, and neural modeling but also leverages recent advances in AI, making UDEs a versatile tool for understanding and controlling neural computation. As illustrated in Figure \ref{UDE for Neuroscience}, UDEs can be applied across diverse domains. Together, these applications demonstrate how UDEs provide a coherent and versatile framework for understanding neural dynamics, guiding interventions, and advancing personalized neuroscience.
\section{Conclusion and Outlook}
This study provides a thorough overview of hybrid mechanistic modeling approaches and potential applications of hybrid mechanistic models for the modeling of neurological diseases. This study comprehensively examines the hybrid modeling approaches based on their configurations, which include parallel, series, and combined parallel–series models, with specific emphasis on residual modeling, Neural Ordinary Differential Equations (Neural ODEs), and solver in the loop approaches. The review discusses how differential programming can be used for mechanistic modeling, which is essential for merging mechanistic models with data driven modeling approaches. Each hybrid model configuration is analyzed for their strengths and weaknesses in terms of modeling incomplete physical dynamics, continuous dynamics based on irregular time series data, and solver based methodologies to improve computational efficiency and stability. Strategies for adaptation based on each type of hybrid model configuration are provided to tackle these issues. Overall, this paper highlights the importance of hybrid modeling in generating accurate, physically feasible, and explainable representations of neurological diseases. Hybrid modeling can overcome many limitations associated with data scarcity, interpretability of models, and scalability. Taken together, all hybrid model configurations have the potential to become an effective approach for modeling neurological diseases.

To highlight the practicality of the above-discussed hybrid model frameworks, it is vital to consider the way the mechanistic model can be made differentiable. The mechanistic tumor growth model was employed as a toy example to demonstrate differentiable programming based seamless integration using automatic differentiation and the adjoint state method for high-dimensional inverse problem solutions, facilitating patient specific calibration from data. From a practical point of view, hybrid modeling has great potential for modeling a wide range of neurological disorders, such as brain tumors, neurodegenerative disorders, cerebrovascular conditions, and traumatic brain injury. All these cases demonstrate the abilities of hybrid models in terms of disease progression forecasting as well as inverse modeling to estimate patient specific parameters. While parallel models prove to be more effective in discovering missing physics in complicated hemodynamical systems, series modeling methods such as Neural ODEs and solver in the loop approaches ensure the mathematical robustness necessary for critical clinical applications like tumor growth forecasting and radiation therapy planning. As for the example of universal differential equations in neuroscience, all these innovations show that hybrid modeling represents not only a novel method but also a paradigm shift in relation to modeling neurological problems for clinical application.

However, despite all these promising developments, several issues have been recognized. One of the major concerns is the heavy computational overhead associated with combining differentiable solvers, particularly in high dimensional and stiff dynamical systems common to brain biomechanics. While hybrid modeling approaches offer an advantage in terms of interpretability over purely data-driven methods, their implementation calls for additional efforts towards further validation and understanding from clinicians' perspective. Given the outlined limitations, one can consider future work that will address the issue of bridging the gap between methodological innovation and clinical application. It will require the use of multimodal patient specific information and the ability to predict future patient trajectories to support clinicians in their decision making process. The emergence of digital twin frameworks for personalized medicine presents a particularly promising direction, wherein hybrid models can function as the computational foundation for simulating disease progression and providing optimal risk averse  treatment strategies. Furthermore, hybrid modeling should be advanced to enable uncertainty quantification in differentiable frameworks to make the modeling more reliable and applicable in the clinical setting. 

In conclusion, differentiable programming based hybrid mechanistic modeling offers a significant step forward by combining the strengths of mechanistic understanding with the flexibility of data driven learning for modeling neurological disorders. While challenges persist, continued advances in computational methods, data availability, and interdisciplinary collaboration are expected to establish hybrid modeling as a central framework in next generation computational neuroscience and precision medicine.
\printbibliography
\end{document}